\documentclass[10pt,twocolumn,letterpaper]{article}
\usepackage[pagenumbers]{cvpr}
\usepackage{tabularx}
\usepackage{multirow}
\usepackage{colortbl}
\usepackage{placeins}
\usepackage{stfloats}
\definecolor{trajblue}{HTML}{245C82}
\definecolor{trajlight}{HTML}{EDF4F8}
\definecolor{trajteal}{HTML}{147D79}
\definecolor{trajorange}{HTML}{AE6032}
\definecolor{refgray}{HTML}{F2F2F2}
\DeclareMathOperator*{\argmin}{arg\,min}

\newcommand{\FIDref}{\ensuremath{\mathrm{FID}}}

\newcommand{\E}{\mathbb{E}}
\newcommand{\TE}{TrajCal-Euler}
\newcommand{\TC}{TrajCal-Correction}
\newcommand{\TU}{TrajCal-UniPC}
\newcommand{\best}[1]{\textbf{#1}}
\newcommand{\second}[1]{\underline{#1}}
\newcommand{\inputrows}[1]{\csname @@input\endcsname{#1}}

\definecolor{cvprblue}{rgb}{0.21,0.49,0.74}
\usepackage[breaklinks,colorlinks,allcolors=cvprblue]{hyperref}
\hypersetup{
  pdftitle={TrajCal: A Calibration Framework for Preserving Reference Fidelity in Few-Step Flow Sampling},
  pdfauthor={Junchi Zhang, Zihan Gu, Jiangcan Jia, Lulu Zhao, Wenbo Su, Hua Zhang},
  pdfsubject={},
  pdfkeywords={}
}
\title{TrajCal: A Calibration Framework for Preserving Reference Fidelity in Few-Step Flow Sampling}
\author{
Junchi Zhang$^{1,}$\thanks{equal contribution}\;,  Zihan Gu$^{2,3,}$\footnotemark[1]\;, Jiangcan Jia$^{4}$,
Lulu Zhao$^{5}$, Wenbo Su$^{5}$, Hua Zhang$^{2,3}$\\[2pt]
{\small $^{1}$Fudan University\quad $^{2}$Institute of Information Engineering, Chinese Academy of Sciences}\\
{\small $^{3}$School of Cyber Security, University of Chinese Academy of Sciences} \;
{\small $^{4}$East China Normal University\quad $^{5}$Alibaba Group}
}
\makeatletter
\patchcmd{\@maketitle}{\vskip .375in}{\vskip 0pt}{}%
  {\PackageError{TrajCal}{Could not adjust the space above the title}{}}
\patchcmd{\@maketitle}{\vspace*{12pt}}{\vspace*{8pt}}{}%
  {\PackageError{TrajCal}{Could not adjust the space below the affiliations}{}}
\patchcmd{\@maketitle}{\vspace*{24pt}}{\vspace*{6pt}}{}%
  {\PackageError{TrajCal}{Could not adjust title spacing}{}}
\apptocmd{\@maketitle}{\noindent
\begin{minipage}{\textwidth}
\centering
\vspace{-14pt}
\captionsetup{type=figure,position=bottom}
\includegraphics[width=\linewidth]{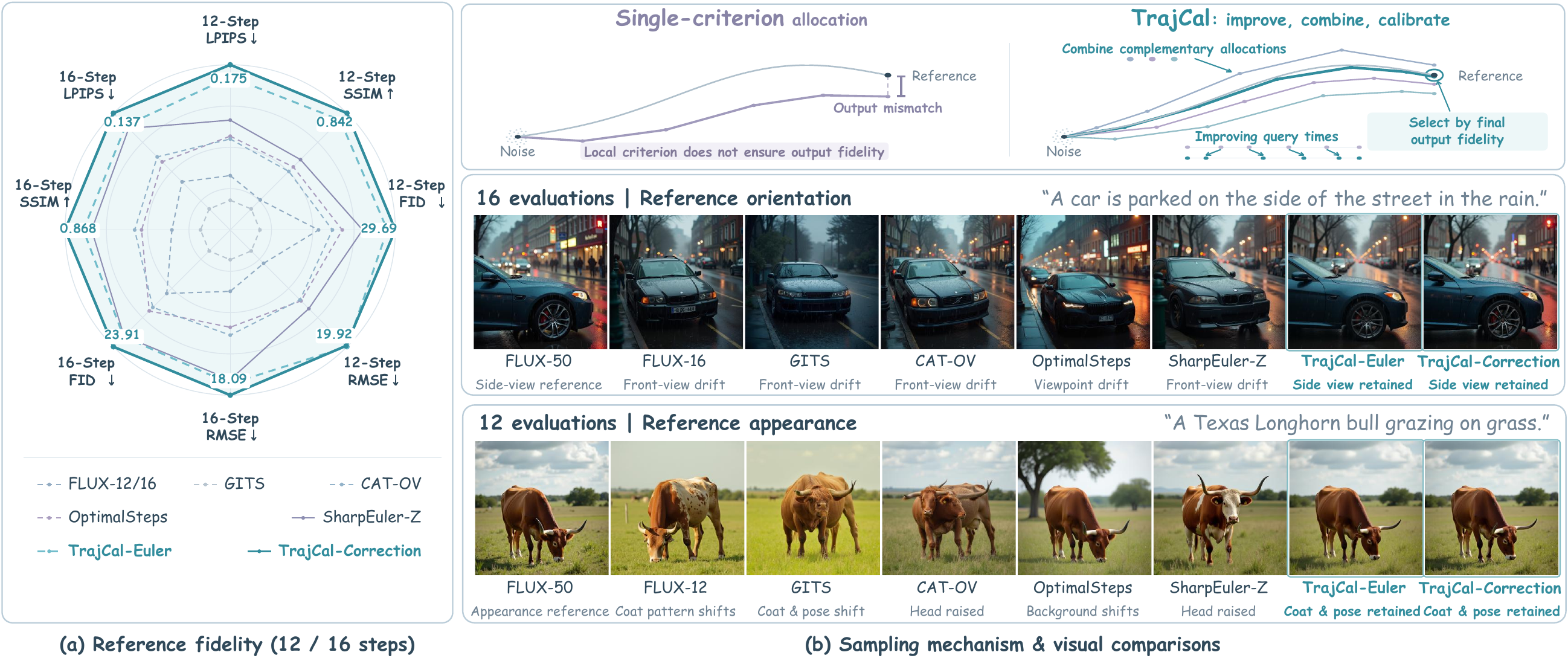}
\caption{\textbf{Reference fidelity with few evaluations.}
(a) RMSE, LPIPS, SSIM, and FID on 500 paired FLUX inputs at 12 and 16 evaluations; larger radii indicate better fidelity, and the labels give \TC{}'s scores.
(b) TrajCal expands seed schedules through mixing and deformation and selects by final-output fidelity.
The examples compare orientation at 16 evaluations and appearance at 12, using the same prompts and base noise as FLUX-50.}
\label{fig:qualitative}
\vspace{0pt}
\end{minipage}
\par}{}%
  {\PackageError{TrajCal}{Could not attach Figure 1 to the title block}{}}
\makeatother
\begin{document}
\maketitle
\begin{abstract}
Flow matching enables high-quality image and video generation, but its iterative sampling process requires repeated evaluations of large neural networks.
Existing few-step methods can distort the output and reduce its agreement with a many-step reference.
We propose \textbf{TrajCal}, a training-free calibration framework for few-step sampling.
Based on scatter search, a well-established metaheuristic for difficult optimization problems, TrajCal starts with diverse seed schedules, deforms and combines these candidates, and selects one fixed schedule by final output fidelity for reuse on new inputs.
For both image and video generation, TrajCal achieves higher reference fidelity than all compared baselines at low evaluation budgets while maintaining competitive visual quality.
The search is effective even with randomly generated seeds, while trajectory-based seeds further improve fidelity.
For cross-style generalization,  photography-calibrated schedules also improve fidelity on unseen image styles.
\end{abstract}

\section{Introduction}
\label{sec:intro}

Flow matching generates images and videos by integrating a learned velocity field that transports noise to data~\cite{lipman2023flowmatching,liu2023rectifiedflow}, supporting models such as FLUX~\cite{bfl2025fluxkontext} and Wan~\cite{wan2025wan}.
This process requires repeated evaluations of large neural networks, making sampling computationally expensive.
This raises the question: can we reduce the number of model evaluations while preserving the output of a many-step sampler under the same conditioning input and base noise?
Existing methods follow two main approaches: training-based methods, such as distillation~\cite{Sabour2025distillation}, reduce sampling cost through additional training, while \emph{training-free} methods keep the pretrained model fixed and approximate the trajectory through selected evaluation times and numerical updates.
In this work, we focus on training-free sampling and aim to improve the reference fidelity of  final outputs under a fixed evaluation budget.

Recent training-free methods use trajectory information and reference
outputs to construct few-step schedules.
GITS~\cite{chen2024trajectory} uses errors along reference trajectories,
OptimalSteps~\cite{pei2025optimalsteps} seeks agreement with many-step
teacher outputs, and SharpEuler~\cite{gupta2026sharpeuler} allocates
evaluations using velocity-change statistics.
These methods provide different strategies for deciding where model evaluations are most useful.
However, the resulting schedules can still leave a substantial fidelity
gap at low evaluation budgets, as shown in Figure~\ref{fig:qualitative}(b): at 16 evaluations, the compared methods shift the car away from the reference side view; at 12 evaluations, they alter the bull's coat pattern, pose, or surrounding scene.

These examples illustrate a limitation of existing methods: their allocation criteria do not ensure agreement with the many-step reference, even when the outputs remain consistent with the prompts.
Similar to~\cite{zhang2026oys}, we formulate schedule design as an optimization problem.
We target fidelity to paired many-step references under a fixed
evaluation budget, and draw on scatter search~\cite{marti2006principles}, a well-established metaheuristic for difficult optimization problems, to construct and explore a useful space of candidate schedules:
starting from a diverse pool of schedules derived from different criteria, we improve and combine their timestep allocations to generate new candidates, and then select a schedule by comparing its final outputs with paired many-step references.

We propose \textbf{TrajCal}, a training-free calibration framework that adapts the above idea to few-step schedule optimization.
TrajCal starts with diverse seed schedules, uses averaging to combine candidates, and applies deformation for further improvement.
Then we screen the expanded candidate pool and select one schedule for each update rule and evaluation budget using a normalized combination of RMSE, LPIPS, and SSIM losses, paired with a many-step reference generated from the same conditioning input and base noise.
The selected schedule is then reused across new inputs without reference generation or further schedule search.
Our main instantiation uses seeds based on weighted dynamic programming~\cite{chen2024trajectory}, acceleration-based allocation~\cite{gupta2026sharpeuler}, and stepsize search~\cite{pei2025optimalsteps}.
The choice is for diversity, and experiments show that randomly generated seeds from distinct shape families also improve reference fidelity, while trajectory-based seeds provide further gains (Section~\ref{sec:ablations}).

For image generation, \TC{} augments Euler updates with a delayed displacement correction.
Consecutive velocity predictions estimate displacement missed over the preceding interval, and a fraction of this estimate is incorporated into the current update without additional model evaluations.
Fig.~\ref{fig:qualitative}(a) shows the improvement of \TC{} across pixel, perceptual, structural, and distributional fidelity measures.
For video generation, \TU{} applies the calibrated schedule while retaining UniPC's predictor--corrector updates and prediction history~\cite{zhao2023unipc}.
The framework therefore supports both a lightweight correction to Euler and an existing higher-order sampler.

Our contributions are as follows:
\begin{itemize}
    \item \textbf{A calibration framework for reference fidelity.}
    We adapt scatter search to construct a diverse candidate space from complementary seeds, expand it through mixing and deformation, and select a reusable schedule for each update rule and budget.

    \item \textbf{Corrected sampling with calibrated schedules.}
    We identify effective update rules for our calibration framework.
    For images, we develop a lightweight delayed displacement correction that requires no additional model evaluations; for videos, we use UniPC.

    \item \textbf{Consistent fidelity improvements across images and videos.}
    TrajCal outperforms the strongest compared baselines in reference fidelity, while maintaining competitive visual quality.
    Photography-calibrated schedules also improve fidelity on unseen styles.
\end{itemize}

\section{Related Work}
\label{sec:related}
\textbf{Timestep selection.}
A sampling schedule distributes a limited evaluation budget along the trajectory.
GITS~\cite{chen2024trajectory} uses dynamic programming guided by reference trajectories;
OptimalSteps~\cite{pei2025optimalsteps} uses dynamic programming to approximate many-step teacher outputs;
SharpEuler~\cite{gupta2026sharpeuler} allocates steps using velocity-change statistics.
Other approaches optimize error bounds or proxies~\cite{sabour2024ays,xue2024steps,zhu2026hso},
or adapt stepsizes during inference~\cite{li2026catflow}.
Some methods jointly search timesteps and update rules~\cite{liu2024usf},
select different solvers based on trajectory curvature~\cite{jo2026sdm},
or fine-tune the velocity network for the selected schedule~\cite{nguyen2024boss}.
OYS~\cite{zhang2026oys} uses Bayesian optimization to search sampling configurations, with HPS or FID as objectives in its image-generation experiments.
TrajCal constructs candidates by mixing and deforming diverse seed schedules,
then selects a reusable schedule by fidelity to paired many-step outputs.

\textbf{Numerical updates.}
DPM-Solver++~\cite{lu2025dpmpp} uses higher-order updates,
while UniPC~\cite{zhao2023unipc} combines prediction and correction
to improve numerical accuracy.
Flow-Solver~\cite{song2024flowsolver} uses past predictions to estimate velocity changes;
SG-prev~\cite{li2026selfguidance} uses the difference between current and previous predictions to guide sampling;
CAB~\cite{roy2026cab} corrects Adams--Bashforth updates using past velocity predictions.
Our image sampler also uses prediction history, with a scaled displacement correction; our video sampler uses UniPC.

\textbf{Sampler learning and reference fidelity.}
Many-step samplers also provide targets for learning time schedules and update coefficients.
DSS~\cite{wang2025dss} and DLMS~\cite{liang2025dlms} use intermediate teacher states; BNS~\cite{shaul2024bns}, S4S-Alt~\cite{frankel2025s4s}, and DyWeight~\cite{zhao2026dyweight} match final teacher outputs.
LD3~\cite{tong2025ld3} and S4S~\cite{frankel2025s4s} relax endpoint matching by allowing bounded perturbations to the initial noise during training.
DLMS~\cite{liang2025dlms} and INDIS~\cite{yuan2026indis} also learn input-dependent sampling parameters.
Faithful Previews~\cite{jia2026faithful} targets alignments with full-step outputs through learned input or trajectory corrections.
TrajCal uses paired output fidelity to select a reusable schedule from mixed and deformed candidates.

\textbf{Caching and step skipping.}
Caching and step skipping reduce full network evaluations by reusing features or extrapolating predictions between refreshes~\cite{liu2025teacache,zhou2025easycache,bajpai2026fastflow}.
Refresh schedules can be guided by approximation errors~\cite{peng2026ertacache,cui2026dpcache,gao2026meancache}, final-output fidelity~\cite{wang2026goldenpath}, or trajectory uncertainty~\cite{lan2026postcache}.
These methods combine computation reuse with decisions about when to recompute; TrajCal focuses on selecting the evaluation times of a few-step sampler.
Appendix~\ref{app:additional} compares visual quality with EasyCache~\cite{zhou2025easycache} at low evaluation budgets.

\section{Method}
\label{sec:method}
In this section, we first define the sampling problem and reference-fidelity objective (Section~\ref{sec:problem}).
Then we describe how scatter search guides candidate construction and expansion (Section~\ref{sec:scatter}).
In Section~\ref{sec:construction},  the complete TrajCal procedure is presented.
\subsection{Problem setup}
\label{sec:problem}
Given a pretrained flow model with fixed parameters $\theta$ and a condition $c$, generation follows the latent trajectory
\begin{equation}
\frac{dx(t)}{dt}=v_\theta(x(t),t;c),\qquad t\in[0,1].
\label{eq:ode}
\end{equation}
Time increases from noise to data. The initial state is obtained from a base noise draw $z$ using the model's initialization convention.
A schedule $T=(t_0,\ldots,t_B)$ contains $B+1$ ordered nodes, with a prescribed initial time $t_0$ and terminal time $t_B=1$.
At step $j$, let
\begin{equation}
h_j=t_{j+1}-t_j,\qquad u_j=v_\theta(x_j,t_j;c),
\label{eq:interval}
\end{equation}
where $x_j$ is the numerical state at $t_j$.
At each query, the update rule uses the available predictions to advance the state.
A $B$-evaluation run queries the velocity at $t_0,\ldots,t_{B-1}$ and ends at $t_B$ without a terminal query.
Decoding the final state gives the generated image or video.

For each condition--noise pair $(c,z)$, the target is the output of the same model under a specified many-step sampler.
For each update rule, we seek a schedule that closely matches this output within a fixed budget $B$.
The model and decoder are held fixed when comparing candidates.
In addition to reference fidelity, visual quality and prompt alignment are also assessed.

\subsection{Scatter search for better schedules}
\label{sec:scatter}
In this part, we describe how scatter search is adapted to construct and expand candidate schedules.
With the pretrained model, update rule, and evaluation budget fixed, schedule construction can be formulated as an optimization problem that maximizes reference fidelity.
To solve this problem, we adapt scatter search~\cite{marti2006principles}, a well-established metaheuristic algorithm for difficult optimization
problems, which can be grouped into four steps: generating diverse starting candidates, improving individual solutions, combining selected solutions to create new candidates, and deciding which candidates to retain.
TrajCal adapts these steps to schedule construction as follows.

We first choose three seed schedules with distinct timestep allocations, and improve each candidate through deformation, adjusting query timing and late-interval allocation while preserving node order and endpoints.
Six deformations are used to explore both signs of three independent directions in deformation-parameter space.
For a smooth fidelity loss $f(x)$ in these parameters, a small move along a direction $d$ satisfies
\begin{equation}
f(x+\epsilon d)=f(x)+\epsilon\nabla f(x)^\top d+O(\epsilon^2).
\label{eq:deformation-local}
\end{equation}
When the gradient is nonzero, at least one of the three directions has a nonzero gradient projection, and one sign makes the first-order term negative.
A sufficiently small move in that direction therefore lowers the loss.

We also combine existing schedules by averaging corresponding time nodes.
For two schedules $T_A$ and $T_B$ with common endpoints, their average $T_M=(T_A+T_B)/2$ preserves node order, endpoints, and budget while exploring an intermediate allocation.
In a second-order expansion of the fidelity loss $g(T)$ around their midpoint, with Hessian $H$ and $\Delta T=T_A-T_B$,
\begin{equation}
g(T_M)\approx\frac{g(T_A)+g(T_B)}{2}-\frac18\Delta T^\top H\Delta T.
\label{eq:mixing-local}
\end{equation}
Positive curvature along the seed difference will give a loss reduction relative to the mean seed loss within this local approximation.
Deformation is also applied to the mixtures to explore further improvements. Detailed derivations are provided in Appendix~\ref{app:analysis}.

Finally, we screen and select candidates by their reference fidelity on calibration inputs.
For efficiency, only one round of scatter search is performed to obtain one schedule for each update rule and budget.
The selected schedule is reused across new inputs without reference generation or further search.

\begin{figure*}[t!]
\centering
\includegraphics[width=\textwidth]{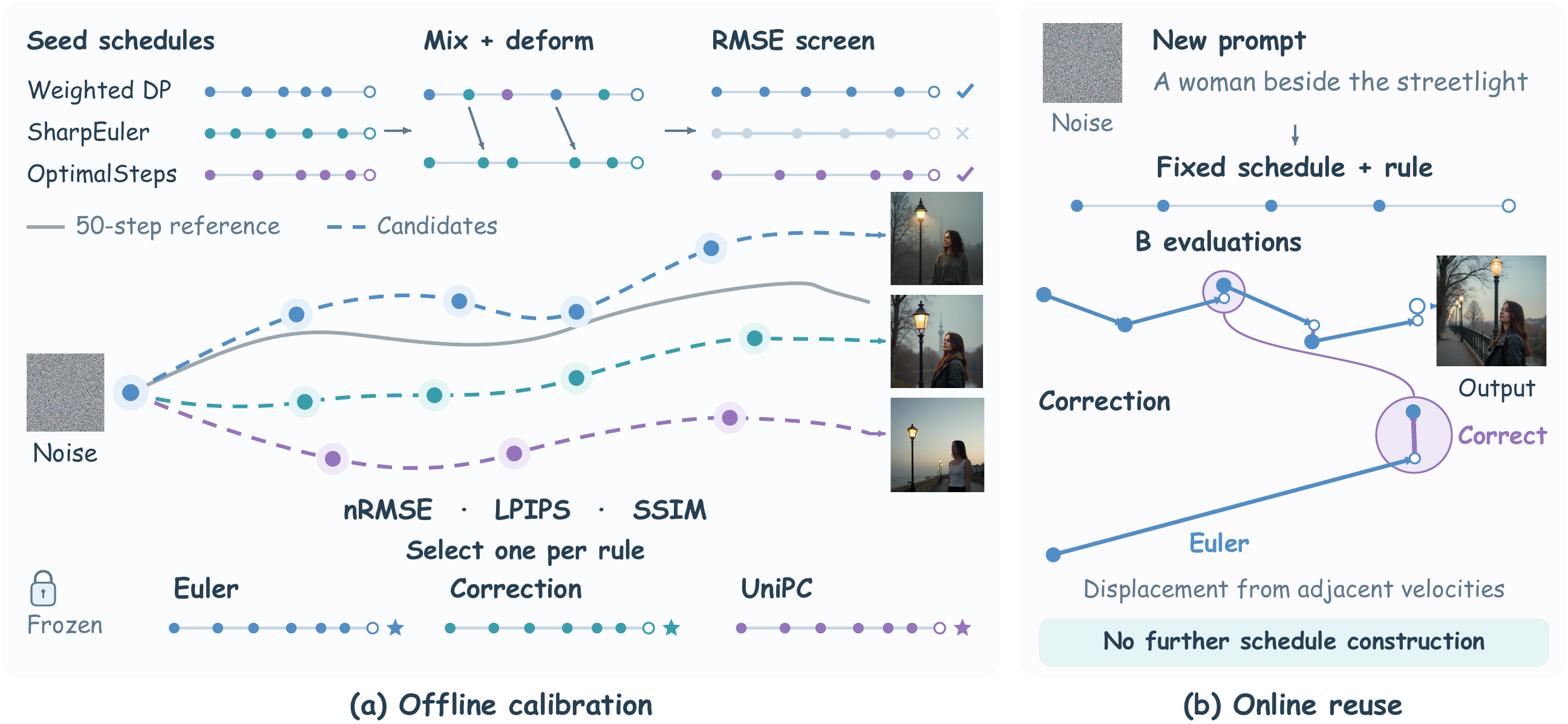}
\vspace{-4pt}
\caption{\textbf{Schedule calibration and reuse in TrajCal.}
(a) Three seed schedules provide distinct time allocations. Mixing and deformation expand the candidate pool, followed by preliminary RMSE screening.
The retained candidates are evaluated against paired many-step references, and a normalized combination of RMSE, LPIPS, and SSIM losses selects one schedule for each update rule and budget.
(b) The selected schedule and update rule are reused on new inputs with $B$ velocity evaluations.
For image sampling, the displacement correction uses consecutive velocity predictions without additional model evaluations.}
\label{fig:framework}
\vspace{-14pt}
\end{figure*}

\subsection{TrajCal: complete process}
\label{sec:construction}
Now we describe the complete TrajCal procedure, which is illustrated in Fig.~\ref{fig:framework}.
Starting from three complementary seed schedules, we align their time ranges and average each pair to obtain six base schedules.
Each contributes its original allocation, two late-interval deformations, and four smooth suffix deformations, yielding a candidate set of 42 schedules before screening.

Using four calibration inputs, an \emph{anchor schedule} is selected by mean output RMSE under the corresponding update rules.
For efficiency, we align each proposal's first $q=\max(2,\lceil B/4\rceil)$ query timesteps to the anchor schedule and affinely rescale the remaining nodes to the endpoint.
The common states and histories allow prefix predictions to be reused for the same input and rule.
Section~\ref{sec:ablations} evaluates the effect of this shared-prefix constraint.
For \TU{}, the native UniPC schedule---its default time grid at budget $B$---is also retained and considered when choosing the anchor schedule.

For preliminary screening, 18 candidates are retained from the expanded pool using eight calibration inputs as follows.
A set of predefined \emph{protected candidates} is initially retained; the score of adding a candidate is determined by the average lowest normalized RMSE achieved by schedules in the enlarged set across each calibration input and update rule.
The candidate giving the lowest resulting mean is added until 18 schedules are retained.
The above process favors complementary candidates rather than simply retaining the 18 lowest individual means.

For a retained schedule, \TC{} deals with image generation, \TU{} deals with video generation, and \TE{} serves as a comparison.

\TE{} uses Euler update, which holds the velocity constant over each interval:
\begin{equation}
x_{j+1}=x_j+h_ju_j.
\label{eq:euler}
\end{equation}
\TC{} adds a displacement correction computed from consecutive velocity predictions.
Over the preceding interval $[t_{j-1},t_j]$, Euler estimates the displacement as $h_{j-1}u_{j-1}$, using only the starting velocity.
Once $u_j$ is computed at $t_j$, we also have a velocity estimate at that interval's endpoint.
Averaging the two predictions gives a trapezoidal estimate, whose difference from the Euler displacement is
\begin{equation}
\delta_j=\frac{h_{j-1}}{2}(u_j-u_{j-1}).
\label{eq:residual}
\end{equation}
This difference estimates displacement missed over the preceding interval.
It becomes available only at $t_j$, and we add a fraction $\eta$ when advancing to $t_{j+1}$:
\begin{equation}
x_{j+1}=x_j+h_ju_j+\eta\delta_j,\qquad 1\leq j<B.
\label{eq:correction}
\end{equation}
The first step uses Euler, since no previous velocity prediction is available.
We fix $\eta=0.5$ and Section~\ref{sec:ablations} compares different correction strengths.
\TU{} uses UniPC~\cite{zhao2023unipc}, retaining its predictor--corrector updates and prediction history while applying the schedule selected by TrajCal.

For each update rule $r$, let $\mathcal T_r$ be the candidate set retained after screening.
We run every $T\in\mathcal T_r$ on the calibration inputs used for final selection, using rule $r$, and compare its final outputs with their many-step references.
For each input $i$ and time schedule $T$, let $R_i(T)$, $P_i(T)$, and $S_i(T)$ denote the RMSE, LPIPS~\cite{zhang2018lpips}, and SSIM~\cite{wang2004ssim}, respectively, with dependence on $r$ suppressed.
For balance, three normalized losses with equal weights are considered simultaneously:
\begin{equation}
L_i(T)=\frac13\left[\frac{R_i(T)}{s_R}
+\frac{P_i(T)}{s_P}+\frac{1-S_i(T)}{s_S}\right].
\label{eq:loss}
\end{equation}
Here, $s_R$, $s_P$, and $s_S$ are normalization scales for RMSE, LPIPS, and $1-\mathrm{SSIM}$, respectively, and definitions are given in Appendix~\ref{app:implementation}.
The final schedule is selected as
\begin{equation}
T_r^*=\argmin_{T\in\mathcal T_r}\frac1n\sum_{i=1}^n L_i(T),
\label{eq:selection}
\end{equation}
where $n$ is the number of calibration inputs used for final selection.
Section~\ref{sec:ablations} compares this choice with selection by RMSE alone.
Generation on a new input follows $T_r^*$ with its corresponding update rule for exactly $B$ velocity evaluations.

\textbf{Notes for choice of seed schedules:}
In this work, we construct seed schedules following GITS, OptimalSteps and SharpEuler, which produce distinct timestep allocations. 
All three methods are included as baselines in the main image experiments.
This choice is useful but not necessary: in Section~\ref{sec:ablations}, we also construct candidates from three families with distinct shapes of timestep size---accelerating growth, slowing growth, and a late peak.
Randomly sampling their growth and curvature parameters, followed by screening on a small calibration set, gives three complementary seed schedules.
Applying TrajCal with these seeds produces a sampler that outperforms all five image baselines in reference fidelity.

Implementation settings are provided in Appendix~\ref{app:implementation}.

\section{Experiments}
\label{sec:experiments}

In this section, we conduct experiments to examine the following questions of TrajCal:

\begin{enumerate}
    \item Can TrajCal improve reference fidelity while maintaining competitive visual quality and prompt alignment?

    \item Do calibrated schedules generalize to unseen styles?

    \item How do the framework's design choices affect the sampling performance?
\end{enumerate}

Section~\ref{sec:setup} introduces the experimental setup.
Sections~\ref{sec:images} and~\ref{sec:video} answer the first question for image and video generation, respectively.
Section~\ref{sec:styles} addresses the second one, and Section~\ref{sec:ablations} addresses the third.

\begin{table*}[t!]
\vspace{-8pt}
\centering
\captionsetup{position=top}
\caption{\textbf{Image generation at 12 and 16 evaluations.}
Results on 500 FLUX.1-dev inputs, with reference fidelity measured against FLUX-50 outputs from the same prompts and base noise.
RMSE, LPIPS, and SSIM are averaged over paired outputs; FID compares the two 500-image sets.
HPSv3 and CLIP assess human preference and prompt alignment.
$B$ denotes the evaluation budget; CAT-OV averages 11.57 and 15.504 evaluations.
Bold and underlined values mark the best and second-best results; shading highlights \TC{}.}
\label{tab:images}
\setlength{\tabcolsep}{3pt}
\renewcommand{\arraystretch}{1.12}
\scalebox{0.8}{%
\begin{tabularx}{\textwidth}{lc*{6}{>{\centering\arraybackslash}X}}
\toprule[1pt]
\multirow{2}{*}{\textbf{Method}} & \multirow{2}{*}{\textbf{Budget}} & \multicolumn{4}{c}{\textbf{Reference fidelity}} & \multicolumn{2}{c}{\textbf{Quality}} \\
\cmidrule(lr){3-6}\cmidrule(l){7-8}
& & \textbf{\boldmath\mbox{RMSE$\downarrow$}} & \textbf{\boldmath\mbox{LPIPS$\downarrow$}} & \textbf{\boldmath\mbox{SSIM$\uparrow$}} & \textbf{\boldmath\mbox{\FIDref{}$\downarrow$}} & \textbf{\boldmath\mbox{HPSv3$\uparrow$}} & \textbf{\boldmath\mbox{CLIP$\uparrow$}} \\
\midrule
FLUX-12 & \multirow{7}{*}{12} & 48.48 & 0.4720 & 0.6403 & 57.24 & 11.86 & \best{0.2633} \\
GITS &  & 52.76 & 0.5380 & 0.6193 & 78.26 & 8.18 & 0.2584 \\
SharpEuler-Z &  & 32.81 & 0.3234 & 0.7338 & 41.43 & \best{13.14} & 0.2620 \\
CAT-OV &  & 35.86 & 0.3739 & 0.7065 & 52.26 & 11.91 & 0.2602 \\
OptimalSteps &  & 35.51 & 0.3664 & 0.7169 & 48.74 & \second{12.52} & \second{0.2621} \\
\TE{} &  & \best{19.60} & \second{0.2079} & \second{0.8258} & \second{32.58} & 12.11 & 0.2607 \\
\rowcolor{trajlight} \TC{} &  & \second{19.92} & \best{0.1750} & \best{0.8417} & \best{29.69} & 12.30 & 0.2593 \\
\specialrule{0.6pt}{0pt}{2pt}
FLUX-16 & \multirow{7}{*}{16} & 44.71 & 0.4218 & 0.6694 & 52.98 & 11.99 & \best{0.2620} \\
GITS &  & 52.83 & 0.5341 & 0.6162 & 76.10 & 8.64 & 0.2583 \\
SharpEuler-Z &  & 22.05 & 0.2006 & 0.8189 & 29.70 & \second{13.09} & 0.2606 \\
CAT-OV &  & 33.48 & 0.3186 & 0.7388 & 45.38 & 12.22 & 0.2610 \\
OptimalSteps &  & 35.50 & 0.3391 & 0.7259 & 43.44 & \best{13.22} & 0.2607 \\
\TE{} &  & \second{21.01} & \second{0.1885} & \second{0.8343} & \second{29.42} & 12.30 & \second{0.2610} \\
\rowcolor{trajlight} \TC{} &  & \best{18.09} & \best{0.1372} & \best{0.8684} & \best{23.91} & 12.43 & 0.2603 \\
\specialrule{0.6pt}{0pt}{2pt}
\rowcolor{refgray} FLUX-50 reference & 50 & --- & --- & --- & --- & 12.45 & 0.2602 \\
\specialrule{1pt}{2pt}{0pt}
\end{tabularx}
}
\end{table*}

\begin{figure*}[t!]
\centering
\vspace{-10pt}
\includegraphics[width=0.8\textwidth]{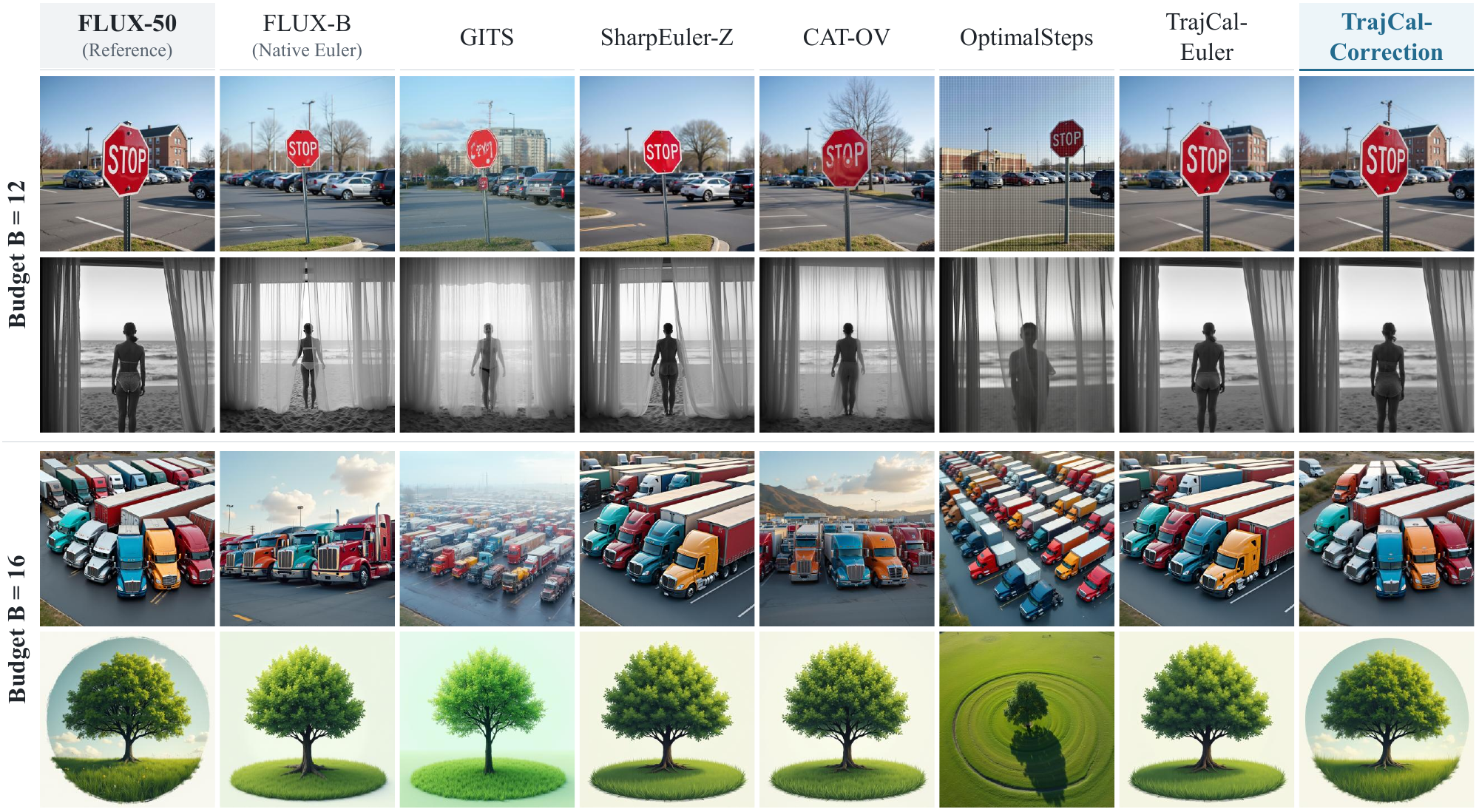}
\caption{\textbf{Visual comparison at 12 and 16 evaluations.}
Each row uses the same prompt and base noise across methods, with the FLUX-50 reference in the first column.
The upper two rows use 12 evaluations and the lower two use 16.
The examples compare scene layout, subject pose, object arrangement, and appearance; \TC{} more closely preserves these features of the reference outputs.}
\label{fig:image_comparison}
\vspace{-8pt}
\end{figure*}

\subsection{Experimental setup}
\label{sec:setup}

We first briefly introduce the models, datasets, compared baselines and evaluation metrics used in our experiments. 

\textbf{Models and data.}
For images, we use FLUX.1-dev~\cite{bfl2024flux}.
The reference, FLUX-50, uses the native schedule and 50 Euler evaluations.
Seed construction uses 64 calibration inputs consisting of COCO~\cite{lin2014coco} caption--noise pairs; final schedule selection uses 128 calibration inputs, including those 64.
We evaluate \TC{} at 12 and 16 evaluations on 500 separate COCO validation pairs, shared across budgets, and include \TE{} as a control.

For video, we use Wan2.2-TI2V-5B~\cite{wan2025wan22}.
The reference, UniPC-50, uses the native UniPC schedule with 50 guided evaluations.
Calibration uses 24 prompt--noise pairs for seed construction and screening and another 24 for final schedule selection; evaluation uses 100 separate pairs.
The three sets are disjoint and drawn from a VBench subset~\cite{huang2024vbench}.
We evaluate \TU{} at 10 and 20 guided evaluations, with \TE{} and \TC{} included for comparison.

\textbf{Comparisons.}
Image baselines are native Euler, GITS~\cite{chen2024trajectory}, SharpEuler-Z~\cite{gupta2026sharpeuler}, CAT-OV~\cite{li2026catflow}, and OptimalSteps~\cite{pei2025optimalsteps}.
Video baselines are native UniPC, StableVS~\cite{yang2026stable}, OptimalSteps, and SharpEuler-Wan.
These methods cover native sampling, trajectory-based allocation, stepsize search, and online adaptation.
SharpEuler-Z is our label for the image variant that allocates all intervals over the full range ending at zero noise; SharpEuler-Wan denotes our adaptation of this construction to Wan2.2 with Euler updates.
CAT-OV is adaptive, averaging 11.57 and 15.504 evaluations at the two image budgets.
Methods share prompts and base noise while retaining their initialization conventions.

\textbf{Measurements.}
Image reference fidelity is measured by RMSE on RGB values in $[0,255]$, LPIPS~\cite{zhang2018lpips}, SSIM~\cite{wang2004ssim}, and FID.
FID compares the generated and FLUX-50 feature distributions on the same inputs: 500 outputs per set in the main study.
HPSv3~\cite{ma2025hpsv3} and CLIP similarity~\cite{radford2021clip} assess human preference and prompt alignment separately.
Video PSNR uses all 81 frames, while LPIPS and SSIM use 16 uniformly sampled frames.
Video quality is assessed by six VBench dimensions: overall consistency (OC), subject consistency (SC), background consistency (BC), motion smoothness (MS), imaging quality (IQ), and dynamic degree (DD).

Appendix~\ref{app:protocol} gives more detailed baseline settings and data protocol.

\subsection{Image generation at 12 and 16 evaluations}
\label{sec:images}
In this part, we evaluate \TC{} on 500 paired FLUX.1-dev inputs at 12 and 16 evaluations, comparing it with five baselines and \TE{} for reference fidelity and image quality.

As shown in Table~\ref{tab:images}, \TC{} improves all four fidelity metrics over every baseline at both budgets.
Compared with SharpEuler-Z, \TC{} improves the three paired metrics simultaneously on 454 of 500 inputs at 12 evaluations and 384 at 16.
For comparison, \TE{} also improves all four measures over the baselines at both budgets, showing the benefit of scatter search with Euler updates unchanged.
At 12 evaluations, it reduces RMSE by 40.2\%, LPIPS by 35.7\%, FID by 21.4\%, and increases SSIM by 12.5\% relative to SharpEuler-Z.
\TC{} reduces LPIPS by a further 15.9\% and 27.2\% at 12 and 16 evaluations, respectively.

The gains also allow better reference fidelity with fewer evaluations.
At 12 evaluations, \TC{} outperforms 16-evaluation SharpEuler-Z, the strongest baseline at that budget, with lower RMSE (19.92 versus 22.05), lower LPIPS (0.1750 versus 0.2006), and higher SSIM (0.8417 versus 0.8189).
It therefore improves all three paired fidelity metrics with 25\% fewer model evaluations, while achieving comparable FID (29.69 versus 29.70).

At both budgets, \TC{} ranks third in HPSv3 among the few-step methods,
while its CLIP similarity remains broadly comparable to the baselines.
These fidelity gains are accompanied by competitive image quality
and prompt alignment.

Figure~\ref{fig:image_comparison} compares the generated images with their paired FLUX-50 references, illustrating the fidelity gains at both evaluation budgets.

\begin{table*}[t!]
\centering
\captionsetup{position=top}
\caption{\textbf{Video generation at 10 and 20 evaluations.}
Results on 100 Wan2.2-TI2V-5B inputs, using paired UniPC-50 references and $B$ guided velocity evaluations.
PSNR uses all 81 frames per video; LPIPS and SSIM use 16 uniformly sampled frames.
The VBench dimensions are overall consistency (OC), subject consistency (SC), background consistency (BC), motion smoothness (MS), imaging quality (IQ), and dynamic degree (DD).
Bold and underlined values mark the best and second-best few-step results within each budget; shading highlights \TU{}.}
\label{tab:video}
\setlength{\tabcolsep}{3pt}
\renewcommand{\arraystretch}{1.12}
\scalebox{0.8}{%
\begin{tabularx}{\textwidth}{lc*{9}{>{\centering\arraybackslash}X}}
\toprule[1pt]
\multirow{2}{*}{\textbf{Method}} & \multirow{2}{*}{\textbf{Budget}} & \multicolumn{3}{c}{\textbf{Reference fidelity}} & \multicolumn{6}{c}{\textbf{VBench dimensions (\boldmath$\times100$)}} \\
\cmidrule(lr){3-5}\cmidrule(l){6-11}
& & \textbf{\mbox{PSNR$\boldsymbol{\uparrow}$}} & \textbf{\mbox{LPIPS$\boldsymbol{\downarrow}$}} & \textbf{\mbox{SSIM$\boldsymbol{\uparrow}$}} & \textbf{\mbox{OC$\boldsymbol{\uparrow}$}} & \textbf{\mbox{SC$\boldsymbol{\uparrow}$}} & \textbf{\mbox{BC$\boldsymbol{\uparrow}$}} & \textbf{\mbox{MS$\boldsymbol{\uparrow}$}} & \textbf{\mbox{IQ$\boldsymbol{\uparrow}$}} & \textbf{\mbox{DD$\boldsymbol{\uparrow}$}} \\
\midrule
UniPC-10 & \multirow{7}{*}{10} & 10.58 & 0.6648 & 0.3994 & 18.61 & 95.81 & 96.72 & 98.11 & 60.11 & 33 \\
StableVS &  & 11.16 & 0.6304 & 0.4167 & 20.01 & 96.28 & 96.92 & 98.04 & 64.34 & 42 \\
OptimalSteps &  & 13.44 & 0.4911 & 0.4985 & \best{21.78} & \second{96.62} & \second{96.98} & \best{98.28} & 65.01 & 49 \\
SharpEuler-Wan &  & 12.26 & 0.5542 & 0.4562 & 21.37 & \best{96.80} & 96.92 & 98.25 & 65.27 & 44 \\
\TE &  & 14.88 & 0.4073 & 0.5392 & \second{21.63} & 96.53 & \best{97.03} & \second{98.27} & 66.92 & 50 \\
\TC &  & \second{15.75} & \second{0.3570} & \second{0.5740} & 21.63 & 96.53 & 96.94 & 98.17 & \second{67.83} & \second{52} \\
\rowcolor{trajlight} \TU & & \best{17.73} & \best{0.2569} & \best{0.6684} & 21.57 & 95.95 & 96.45 & 97.86 & \best{68.32} & \best{56} \\
\specialrule{0.6pt}{0pt}{2pt}
UniPC-20 & \multirow{7}{*}{20} & 11.65 & 0.5626 & 0.4379 & 21.22 & 95.82 & 96.75 & 97.97 & 66.18 & 48 \\
StableVS &  & 12.89 & 0.4856 & 0.4870 & 21.64 & 96.25 & \second{96.78} & 98.04 & 68.22 & 51 \\
OptimalSteps &  & 15.06 & 0.3788 & 0.5639 & 21.84 & \best{96.32} & 96.72 & \second{98.09} & 69.00 & 53 \\
SharpEuler-Wan &  & 13.97 & 0.4309 & 0.5171 & 21.73 & \second{96.32} & \best{96.93} & \best{98.10} & 67.85 & 54 \\
\TE &  & \second{15.15} & \second{0.3698} & 0.5618 & \best{22.17} & 96.17 & 96.74 & 98.04 & \best{69.40} & 54 \\
\TC &  & 14.86 & 0.3763 & \second{0.5639} & \second{21.86} & 95.87 & 96.51 & 98.05 & \second{69.27} & \best{58} \\
\rowcolor{trajlight} \TU & & \best{24.81} & \best{0.0810} & \best{0.8745} & 21.71 & 95.91 & 96.63 & 97.91 & 68.62 & \second{56} \\
\specialrule{0.6pt}{0pt}{2pt}
\rowcolor{refgray} UniPC-50 reference & 50 & --- & --- & --- & 21.68 & 95.69 & 96.55 & 97.90 & 68.52 & 54 \\
\specialrule{1pt}{2pt}{0pt}
\end{tabularx}%
}
\vspace{-14pt}
\end{table*}

\subsection{Video generation at 10 and 20 evaluations}
\label{sec:video}
In this part, we evaluate \TU{} on 100 paired Wan2.2-TI2V-5B inputs at 10 and 20 evaluations, comparing it with four baselines and \TE{} and \TC{} for reference fidelity and video quality.

\TU{} improves PSNR, LPIPS, and SSIM over every baseline in Table~\ref{tab:video} at both video budgets.
Compared with OptimalSteps, the strongest baseline in these measures, it raises PSNR by 4.29 dB, reduces LPIPS by 47.7\%, and increases SSIM from 0.4985 to 0.6684 at 10 evaluations.
The advantage grows at 20 evaluations: PSNR rises from 15.06 to 24.81 dB, LPIPS falls by 78.6\%, and SSIM increases from 0.5639 to 0.8745.
These gains are consistent across individual inputs: \TU{} improves all three measures simultaneously over OptimalSteps on all 100 videos at each budget.

The improvement also enables a substantial reduction in the evaluation budget.
Ten-evaluation \TU{} outperforms the strongest 20-evaluation baseline in PSNR (17.73 versus 15.06 dB), LPIPS (0.2569 versus 0.3788), and SSIM (0.6684 versus 0.5639).
It therefore achieves better reference fidelity with half as many guided evaluations.
All three metrics improve simultaneously on 89 of the 100 videos in this cross-budget comparison (Appendix~\ref{app:additional}).

For comparison of different update rules, \TC{} improves all three fidelity measures over \TE{} at 10 evaluations, and \TU{} improves them further.
At 20 evaluations, \TE{} and \TC{} give similar fidelity, while \TU{} has a substantially larger advantage.

VBench scores indicate that the fidelity improvements accompany comparable video quality.
\TU{} has imaging-quality scores of 68.32 and 68.62 at the two budgets, close to the reference's 68.52, and its overall consistency also remains close to the reference.
Subject consistency, background consistency, and motion smoothness are similar to the reference, although some baselines score higher on these dimensions.
Dynamic degree is 56 at both budgets, compared with 54 for the reference.

\subsection{Cross-style generalization}\label{sec:styles}
\begin{table}[t!]
\centering
\captionsetup{position=top}
\caption{\textbf{Cross-style generalization.}
Schedules calibrated on photography are reused unchanged on 30 scenes in four styles: photography, watercolor, anime, and flat-vector art.
FID is computed from all 120 generated images and their references; the other three metrics are averaged over these outputs.
Bold, underlining, and shading follow Table~\ref{tab:images}.}
\label{tab:styles}
\footnotesize
\setlength{\tabcolsep}{2pt}
\renewcommand{\arraystretch}{1.12}
\scalebox{1}{%
\begin{tabularx}{\columnwidth}{lc*{4}{>{\centering\arraybackslash}X}}
\toprule[1pt]
\multirow{2}{*}{\textbf{Method}} & \multirow{2}{*}{\textbf{Budget}} & \multicolumn{4}{c}{\textbf{Reference fidelity}} \\
\cmidrule(l){3-6}
& & \textbf{\boldmath\mbox{RMSE$\downarrow$}} & \textbf{\boldmath\mbox{LPIPS$\downarrow$}} & \textbf{\boldmath\mbox{SSIM$\uparrow$}} & \textbf{\boldmath\mbox{\FIDref{}$\downarrow$}} \\
\midrule
FLUX-12 & \multirow{7}{*}{12} & 56.05 & 0.4813 & 0.6879 & 125.27 \\
GITS &  & 59.57 & 0.5402 & 0.6844 & 169.76 \\
SharpEuler-Z &  & 41.27 & 0.3374 & 0.7437 & 82.38 \\
CAT-OV &  & 40.97 & 0.3557 & 0.7326 & 91.34 \\
OptimalSteps &  & 42.50 & 0.3787 & 0.7232 & 91.40 \\
\TE{} &  & \best{23.18} & \second{0.2077} & \second{0.8192} & \second{59.78} \\
\rowcolor{trajlight} \TC{} &  & \second{24.45} & \best{0.1850} & \best{0.8280} & \best{54.50} \\
\specialrule{0.6pt}{0pt}{2pt}
FLUX-16 & \multirow{7}{*}{16} & 52.92 & 0.4416 & 0.7016 & 112.34 \\
GITS &  & 59.61 & 0.5372 & 0.6812 & 165.19 \\
SharpEuler-Z &  & 26.84 & 0.2017 & 0.8202 & 56.59 \\
CAT-OV &  & 39.66 & 0.3198 & 0.7524 & 81.80 \\
OptimalSteps &  & 43.52 & 0.3669 & 0.7239 & 84.33 \\
\TE{} &  & \best{20.97} & \second{0.1594} & \second{0.8534} & \second{47.06} \\
\rowcolor{trajlight} \TC{} &  & \second{22.25} & \best{0.1446} & \best{0.8580} & \best{46.26} \\
\specialrule{1pt}{0pt}{0pt}
\end{tabularx}%
}
\vspace{-12pt}
\end{table}

In this part, we test whether schedules selected using photography prompts also improve fidelity in other image styles.
Calibration uses only photography inputs.
We apply the selected schedules unchanged to 30 scenes excluded from calibration, generating each scene in four styles: photography, watercolor, anime, and flat-vector art.
When generating different styles of the same scene, we keep the scene description and initial noise fixed and change only the style prefix in the prompt.
Each output is compared with the FLUX-50 reference generated from the same styled prompt and noise.
Table~\ref{tab:styles} reports four reference-fidelity metrics across all four styles, using 120 outputs per method.
The three paired metrics are averaged over these outputs; FID is computed on the pooled outputs and references.

Across the four styles, \TC{} improves all four fidelity metrics over the baselines at both budgets.
At 12 evaluations, \TC{} achieves mean LPIPS and SSIM of 0.1850 and 0.8280, compared with 0.3374 and 0.7437 for the best baseline, and reduces FID from 82.38 to 54.50.
At both budgets, \TC{} gives better LPIPS, SSIM, and FID than \TE{}, with slightly higher RMSE.
On the three unseen styles, \TC{} lowers mean LPIPS from 0.3392 to 0.1774 at 12 evaluations and from 0.2009 to 0.1393 at 16 relative to SharpEuler-Z.
It also improves RMSE, SSIM, and pooled FID over every baseline on these unseen styles at both budgets.
Appendix~\ref{app:additional} reports the unseen-style results and all six metrics separately for each style.

\subsection{Controlled comparisons and ablations}
\label{sec:ablations}
In this part, we conduct several controlled comparisons and ablations to assess how TrajCal's design choices affect reference fidelity.
Unless otherwise stated, the controls below use 100 image evaluation inputs excluded from calibration.

\textbf{Seed construction.}
We randomly sample 12 candidate schedules from three shape families and select three seeds using eight calibration inputs. Applying TrajCal to these seeds already yields strong reference fidelity.
On 200 FLUX inputs at 12 evaluations, both \TE{} and \TC{} outperform all five image baselines on all four fidelity metrics (Appendix~\ref{app:random-seeds}).
Compared with SharpEuler-Z, \TC{} with random seeds reduces RMSE from 33.70 to 24.63, LPIPS from 0.3286 to 0.2271, and FID from 58.80 to 42.95, while increasing SSIM from 0.7275 to 0.7941.
Replacing these seeds with our three trajectory-based schedules, while keeping candidate expansion and selection unchanged, further improves all four metrics for both update rules.
For \TC{}, RMSE and LPIPS decrease by a further 19.8\% and 23.5\%, respectively.
These results show that candidate search is effective with randomly constructed seeds and that trajectory information provides stronger starting points.

\textbf{Selection objective.}
We compare the combined loss $L_i(T)$ with RMSE-only selection over the same retained candidates.
For \TE{} at 12 evaluations, it lowers LPIPS from 0.2171 to 0.2061 and also improves RMSE and SSIM on evaluation inputs (Appendix~\ref{app:design}).
For \TC{}, both objectives select the same schedule.

\textbf{Correction strength.}
We reselect a schedule for each correction strength from the same 18 candidates.
At 12 evaluations, $\eta=0.25$ and $0.5$ both outperform $\eta=1$ in all three fidelity measures.
The default $\eta=0.5$ gives slightly lower LPIPS than $\eta=0.25$, while $\eta=0.25$ gives lower RMSE and higher SSIM (Appendix~\ref{app:design}).

\textbf{Shared prefix.}
We remove prefix alignment and repeat screening and selection (Appendix~\ref{app:design}).
At 12 evaluations, \TC{}'s LPIPS changes from 0.1779 to 0.1803, with similarly small changes in RMSE and SSIM.
Shared-prefix reuse reduces model calls for final candidate evaluation from 27,648 to 18,944 per rule, a 31.5\% saving for this stage alone (Appendix~\ref{app:efficiency}).

\textbf{Additional budgets.}
We extend the evaluation beyond the main 12- and 16-evaluation settings to budgets of 8 and 20 on 200 paired FLUX inputs (Appendix~\ref{app:additional}).
\TC{} improves all four fidelity measures over all five baselines at both budgets.
Compared with best baselines, \TC{} reduces RMSE by 12.1\% and LPIPS by 14.8\% at 8 evaluations, and reduces LPIPS by 26.1\% and FID by 17.8\% at 20 evaluations.
\section{Conclusion}
\label{sec:conclusion}
TrajCal chooses diverse seed schedules as the starting point for a broader search guided by the reference fidelity of final outputs.
Mixing and deforming complementary seeds expands the candidate space, and calibration selects one reusable schedule for each update rule using forward evaluations only.
The random-seed experiment shows that this procedure can be effective without recorded trajectories for initial node placement; trajectory-based seeds further improve fidelity in the same comparison.
The framework improves reference fidelity in image and video generation; for cross-style generalization, photography-selected schedules also transfer to unseen styles.

\textbf{Limitations and future work.}
TrajCal requires time-consuming candidate evaluations during calibration.
In addition, the selected schedule is shared across inputs and does not adapt its time allocation to new inputs.
Finally, the current method depends on some empirical design choices. A more systematic framework for few-step sampling remains to be developed.

\FloatBarrier
\label{page:mainend}
\section*{LLM Usage Statement}
OpenAI ChatGPT 6 Astra assisted with discussions of research ideas, manuscript writing, literature organization, code development, and analysis of experimental results.
The authors take full responsibility for the final manuscript, code, results, and conclusions.

{\small
\bibliographystyle{ieeenat_fullname}
\bibliography{references}
}
\clearpage
\appendix
\maketitlesupplementary
This supplement follows the notation and experimental settings of the main paper.
Appendix~\ref{app:implementation} specifies the calibration algorithm and its implementation.
Appendix~\ref{app:analysis} analyzes the properties of mixing, deformation, and displacement correction.
Appendix~\ref{app:protocol} specifies the experimental protocol, including data use, metrics, and baseline settings.
Appendix~\ref{app:random-seeds} compares random and trajectory-based seed construction.
Appendix~\ref{app:design} reports the selection-objective, correction-strength, and shared-prefix controls and update-rule comparison.
Appendix~\ref{app:additional} presents cross-style results, additional budgets, paired statistics, and visual comparisons for video generation and caching.
Appendix~\ref{app:efficiency} reports generation time and shared-prefix call savings.
\raggedbottom
\makeatletter
\setlength{\@fptop}{0pt}
\setlength{\@dblfptop}{0pt}
\setlength{\@fpsep}{12pt}
\setlength{\@dblfpsep}{12pt}
\makeatother
\newcommand{\apptablesetup}{\small\setlength{\tabcolsep}{3.5pt}\renewcommand{\arraystretch}{1.12}}
\newcommand{\apphead}[1]{\textbf{\boldmath\mbox{#1}}}
\section{Search procedure and implementation}
\label{app:implementation}

This section specifies seed construction, candidate expansion, and schedule selection for Section~\ref{sec:construction}.
Section~\ref{app:construction} gives the three trajectory-based seed constructions.
Section~\ref{app:candidates} then describes mixing and deformation, shared-prefix alignment, preliminary screening, and the final selection of one reusable schedule for each update rule.
Algorithm~\ref{alg:calibration} summarizes the procedure.
Numerical settings are given in Appendix~\ref{app:protocol}.

We use the increasing time variable $t$, with interval lengths $h_j=t_{j+1}-t_j$ and budget $B$ as in the main paper.
There are $B$ velocity queries and $B+1$ nodes; the endpoint $t_B=1$ is not queried.
Let $\mathcal D_c$ and $\mathcal D_s$ denote the calibration inputs used for seed construction and preliminary screening, and for final schedule selection, respectively.
Their sizes and overlap are given in Appendix~\ref{app:protocol}.
\TE{} and \TC{} share the screened pool but select a schedule separately; \TU{} uses its own UniPC-based screening and selection.

\newcounter{trajcalalgorithm}
\begin{table}[t!]
\centering
\begin{minipage}{\linewidth}
\hrule
\vspace{4pt}
\refstepcounter{trajcalalgorithm}\label{alg:calibration}%
\textbf{Algorithm~\thetrajcalalgorithm: Schedule calibration}\\[3pt]
\textbf{Input:} Frozen model, budget $B$, update rules, calibration inputs $\mathcal D_c$ and $\mathcal D_s$, and their many-step references.\\
\textbf{Output:} One reusable schedule $T_r^*$ for each update rule $r$.
\begin{list}{\arabic{enumi}.}{\usecounter{enumi}\setlength{\leftmargin}{1.65em}\setlength{\labelwidth}{1.15em}\setlength{\labelsep}{0.5em}\setlength{\topsep}{4pt}\setlength{\parsep}{0pt}\setlength{\itemsep}{1pt}}
\item Obtain three seeds from trajectory-based constructions (Appendix~\ref{app:construction}) or random shapes (Appendix~\ref{app:random-seeds}). Align their endpoints.
\item Add the three pairwise node averages. Apply two late-interval and four suffix deformations to each of the six base schedules, retaining the bases: 42 proposals in total.
\item Process Euler and correction jointly, and UniPC separately. For UniPC, also include its native schedule.
\item Use four inputs from $\mathcal D_c$ to choose an anchor among the base schedules by mean output RMSE under the relevant rules. Include the native schedule in the UniPC anchor comparison.
\item Align the first $q=\max(2,\lceil B/4\rceil)$ query nodes to the anchor and rescale the remaining nodes. Keep the native UniPC grid unchanged. Merge duplicate schedules.
\item Use eight inputs from $\mathcal D_c$ to screen the pool. Keep the protected candidates and greedily add complementary schedules until 18 remain.
\item Evaluate the retained schedules on $\mathcal D_s$ under each rule $r$. Normalize RMSE, LPIPS, and $1-\mathrm{SSIM}$ and select the schedule with the lowest mean combined loss.
\end{list}
\vspace{2pt}\hrule
\end{minipage}
\end{table}

\subsection{Trajectory-based seeds}
\label{app:construction}
The three seeds use complementary trajectory-based schedules: weighted dynamic programming, acceleration-based allocation, and stepsize search, following GITS~\cite{chen2024trajectory}, SharpEuler~\cite{gupta2026sharpeuler}, and OptimalSteps~\cite{pei2025optimalsteps}, respectively.

\textbf{Weighted dynamic programming.}
For each calibration input $i\in\mathcal D_c$, let $x_{i,p}$ and $u_{i,p}$ denote the trajectory state and velocity at time $a_p$, where $a_0<\cdots<a_M$ and $\Delta a_p=a_{p+1}-a_p$.
Images use $a_p=p/128$ with an additional terminal velocity query; Wan uses its native 50-interval grid with
\begin{equation}
u_{i,M}=u_{i,M-1}+\frac{\Delta a_{M-1}}{\Delta a_{M-2}}
(u_{i,M-1}-u_{i,M-2}).
\end{equation}
Define the edge cost by
\begin{align}
d_i(p,q)&=\sum_{k=p}^{q-1}\Delta a_k
\left(\frac{u_{i,k}+u_{i,k+1}}2-u_{i,p}\right),\nonumber\\
C(p,q)&=w_p^2\sqrt{\E_{i\in\mathcal D_c}\|d_i(p,q)\|_{\rm RMS}^2},
\label{app:eq:dp}
\end{align}
where $\|v\|_{\rm RMS}^2$ is the mean of squared latent coordinates.
Let $X_{p,b}$ and $U_{p,b}$ be spatial Fourier coefficients of states and velocities centered across $\mathcal D_c$, restricted to radial bands with boundaries $0,1/16,1/8,1/4,\infty$ (cycles per latent-grid sample).
With sums over inputs and band coefficients, set
\begin{align}
\lambda_{p,b}&=\frac{\sum\operatorname{Re}(\overline X_{p,b}U_{p,b})}
{\max(\sum|X_{p,b}|^2,10^{-20})},\quad
\lambda_p=\max(0,\max_b\lambda_{p,b}),\nonumber\\
w_p&=\min\left\{32,\exp\left(\sum_{k=p}^{M-1}
\frac{\lambda_k+\lambda_{k+1}}2\Delta a_k\right)\right\}.
\end{align}
The overbar denotes complex conjugation.
Dynamic programming returns the seed $T^{\rm DP}=(a_{p_j^*})_{j=0}^B$, where
\begin{equation}
(p_0^*,\ldots,p_B^*)\in
\arg\min_{0=p_0<\cdots<p_B=M}\sum_{j=0}^{B-1}C(p_j,p_{j+1}).
\end{equation}

\textbf{Acceleration-based allocation.}
On native 50-interval Euler trajectories for $i\in\mathcal D_c$, let $\tau_j$ be the query times. With $\epsilon=10^{-12}$, define for $j=0,\ldots,48$,
\begin{align}
A_{i,j}&=\frac{\|u_i(\tau_{j+1})-u_i(\tau_j)\|_{\rm RMS}}
{\tau_{j+1}-\tau_j},\nonumber\\
\bar A_j&=\E_{i\in\mathcal D_c}\!\left[
\frac{A_{i,j}}{\max(\sum_\ell A_{i,\ell},\epsilon)}\right],\nonumber\\
\phi&\propto\mathcal G_1\!\left(\sqrt{\bar A+\epsilon}\right),
\qquad \sum_j\phi_j=1.
\end{align}
Here, $\mathcal G_1$ denotes Gaussian smoothing along the index $j$, with a standard deviation of 1.
For images, let $t_\star$ be the last query time of the native $B$-step FLUX schedule and allocate $K=B-1$ intervals on $[0,t_\star]$; for Wan, use $t_\star=1$ and $K=B$.
Form interpolation pairs $(c_j,m_j)_{j=0}^{49}$ using
\begin{align}
m_j&=\min\!\left\{\frac{\tau_j+\tau_{j+1}}2,t_\star\right\},
\quad j=0,\ldots,48,\nonumber\\
c_j&=\sum_{\ell=0}^{j-1}\phi_\ell,\quad j=1,\ldots,49,
\end{align}
with $c_0=0$ and $m_{49}=t_\star$.
Let $Q$ be the piecewise-linear interpolant through these pairs and set
\begin{equation}
t_0=0,\qquad t_k=Q(k/K),\quad k=1,\ldots,K.
\end{equation}
For images, append $t_B=1$, preserving the native final interval $[t_\star,1]$.

\textbf{Stepsize search.}
For the main image experiments, we use the FLUX schedule supplied by OptimalSteps.
We resample it to $B+1$ noise levels using the authors' log-linear interpolation, set the final level to zero, and convert to $t=1-\sigma$, where $\sigma$ denotes the noise level.
For Wan2.2-TI2V-5B, we use the OptimalSteps Wan schedule at budget $B$, with the same time conversion.

Before mixing, we linearly rescale the acceleration-based and stepsize-search schedules to the DP time interval \([a_0,1]\). In our experiments, \(a_0=0\) for images and \(a_0\approx 0.0002\) for videos.

\subsection{Expansion, screening, and selection}
\label{app:candidates}
\begin{figure*}[t!]
\centering
\includegraphics[width=0.8\textwidth]{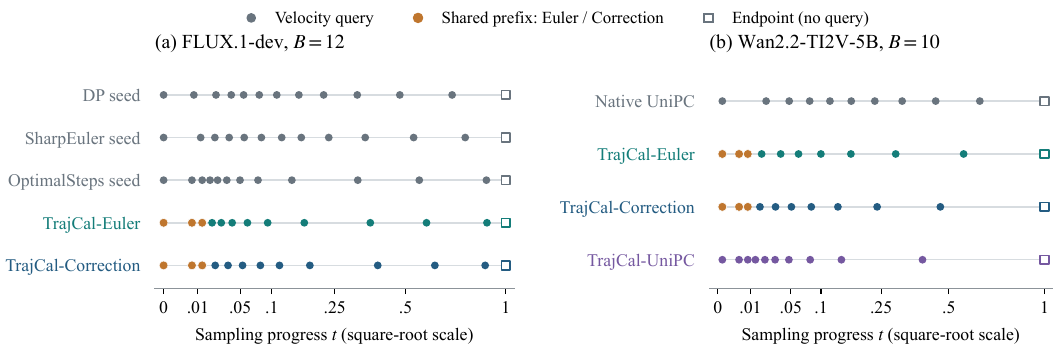}
\caption{\textbf{Seed schedules and schedules selected by TrajCal.} Each row shows the $B$ model evaluation times and the final endpoint, where no evaluation is made.
Left: the three FLUX seeds and the schedules selected for \TE{} and \TC{} at $B=12$. Right: native UniPC and the schedules selected for \TE{}, \TC{}, and \TU{} on Wan at $B=10$.
Orange marks the first three query times shared by \TE{} and \TC{}; \TU{} is calibrated separately.
The horizontal axis uses a square-root scale to make early times easier to see; tick labels give the actual sampling time $t$.
The FLUX seeds are shown before prefix alignment.}
\label{fig:selected_schedules}
\end{figure*}

The same expansion and selection rules apply after either trajectory-based or random seed construction.
Table~\ref{tab:search_settings} collects the fixed parameters.
Figure~\ref{fig:selected_schedules} shows the construction seeds and selected schedules.
\begin{table}[t!]
\centering
\apptablesetup
\captionsetup{position=top}
\caption{\textbf{Search settings.} Counts are for one model at one evaluation budget. \TU{} also includes the native UniPC schedule.}
\label{tab:search_settings}
\begin{tabularx}{\columnwidth}{@{}>{\raggedright\arraybackslash}p{0.43\columnwidth}>{\raggedright\arraybackslash}X@{}}
\toprule[1pt]
\apphead{Parameter} & \apphead{Value}\\
\midrule
Seed schedules & 3\\
Pairwise averages & 3, with weight $1/2$\\
Candidates per base schedule & 7: original schedule + 6 deformations\\
Late-interval deformations & 2, with $\alpha\in\{-0.35,0.35\}$\\
Suffix deformations & 4, with $(a,b)\in\{-0.15,0.15\}^2$\\
Candidates after expansion & 42; 43 for \TU{}, before merging duplicates\\
Anchor selection & 4 calibration inputs from $\mathcal D_c$\\
Shared-prefix length & $q=\max(2,\lceil B/4\rceil)$\\
Screening & 8 calibration inputs from $\mathcal D_c$\\
Candidates after screening & 18\\
Selection objective & Mean of normalized RMSE, LPIPS, and $1-\mathrm{SSIM}$\\
Correction strength & $\eta=0.5$\\
\bottomrule[1pt]
\end{tabularx}
\end{table}

\textbf{Candidate expansion.}
The three seeds and their pairwise node averages give six base schedules.
Each base contributes seven candidates: the base itself, two late-interval deformations, and four suffix deformations.
For the late-interval deformation, set $\ell=\max(2,\lceil0.75B\rceil)$ and $\xi_j=2(j-\ell+1)/(B-\ell)-1$ for $j=\ell-1,\ldots,B-1$.
With $\alpha\in\{-0.35,0.35\}$, replace the affected intervals by
\begin{equation}
\widehat h_j=(1-t_{\ell-1})
\frac{h_j e^{\alpha\xi_j}}{\sum_{k=\ell-1}^{B-1}h_k e^{\alpha\xi_k}}.
\label{app:eq:late-warp}
\end{equation}
Earlier nodes remain fixed; cumulative sums of $\widehat h_j$ give the new suffix.
For the suffix deformation, map $[t_1,1]$ to $y\in[0,1]$, apply
\begin{equation}
W(y)=y+a y(1-y)+b y(1-y)(2y-1),
\label{app:eq:warp}
\end{equation}
and map back, with $(a,b)\in\{-0.15,0.15\}^2$.
This leaves the first interval and terminal endpoint fixed.
After suffix deformation, we return the image DP nodes to the $1/128$ grid.
Grid indices within $10^{-10}$ of an integer are set to that integer; all others are rounded down.
We then adjust the indices to keep adjacent nodes at least one cell apart and leave room for the remaining nodes.
Other image bases, all Wan bases, and random-shape bases retain continuous times.
These operations produce 42 candidates before prefix alignment and deduplication.
For Wan, we retain the native schedule as a candidate for UniPC selection, producing 43 candidates in total.

\textbf{Anchor and shared prefix.}
The first four inputs in $\mathcal D_c$ select an anchor among the six base schedules by mean RMSE over Euler and correction updates.
UniPC selects its anchor separately, comparing the six bases and its native schedule under UniPC updates.
Let $A$ denote the anchor and $q=\max(2,\lceil B/4\rceil)$.
For each proposal $T$, the aligned schedule is
\begin{equation}
\widetilde t_j=\begin{cases}
A_j,&j<q,\\
A_{q-1}+(1-A_{q-1})\dfrac{t_j-t_{q-1}}{1-t_{q-1}},&j\geq q.
\end{cases}
\end{equation}
The $q$ shared positions include $t_0$.
Anchor ties are resolved by candidate name.
Candidates with identical prefixes can reuse network predictions during calibration.
The native UniPC schedule keeps its original nodes.

\textbf{Screening.}
After merging duplicates, we evaluate the pool on the first eight inputs in $\mathcal D_c$.
For Euler and correction, the protected set contains the six bases and the DP suffix variants with $(a,b)=(-0.15,0.15)$ and $(0.15,0.15)$, with aliases resolved after deduplication.
For UniPC, it contains the six bases and the native schedule.
For joint Euler/correction screening, each rule's RMSE matrix is divided by its lowest candidate mean, with a $10^{-9}$ floor.
UniPC screening uses its raw RMSE matrix.
Writing these screening errors as $\widetilde R_{i,r}(T)$, score a retained set $\mathcal S$ by
\begin{equation}
J(\mathcal S)=\frac{1}{m|G|}\sum_{i=1}^{m}\sum_{r\in G}
\min_{T\in\mathcal S}\widetilde R_{i,r}(T),
\label{app:eq:screening}
\end{equation}
where $m=8$ and $G$ contains either Euler and correction or UniPC alone.
Starting with the protected set, add the candidate that minimizes $J(\mathcal S\cup\{T\})$ until 18 remain.
Taking the minimum separately for each input and rule rewards candidates that improve the retained set on different inputs.
Ties favor the largest minimum Euclidean distance between log-interval vectors, then candidate name.

\textbf{Final selection.}
Every retained candidate is evaluated on $\mathcal D_s$ under the corresponding update rule $r$.
We write the three fidelity losses as
$\ell_{i,R}(T)=R_i(T)$, $\ell_{i,P}(T)=P_i(T)$, and $\ell_{i,S}(T)=1-S_i(T)$.
For $k\in\{R,P,S\}$, define
\begin{align}
\bar\ell_{i,k}&=\frac{1}{|\mathcal T_r|}\sum_{T\in\mathcal T_r}\ell_{i,k}(T),\nonumber\\
s_k&=\max\!\left\{\sqrt{\frac{\sum_{i,T}(\ell_{i,k}(T)-\bar\ell_{i,k})^2}{n|\mathcal T_r|}},10^{-8}\right\}.
\label{app:eq:scales}
\end{align}
Here $\mathcal T_r$ is the retained pool for rule $r$, $n=|\mathcal D_s|$, and the sums run over $i=1,\ldots,n$ and $T\in\mathcal T_r$.
We compute the scales separately for each rule and select the schedule with the lowest mean combined loss:
\begin{align}
L_i(T)&=\frac13\left[\frac{R_i(T)}{s_R}
+\frac{P_i(T)}{s_P}+\frac{1-S_i(T)}{s_S}\right],
\label{app:eq:loss}\\
T_r^*&=\argmin_{T\in\mathcal T_r}\frac1n\sum_{i=1}^n L_i(T).
\label{app:eq:selection}
\end{align}

\FloatBarrier
\section{Analysis of the search and updates}
\label{app:analysis}

We derive the feasibility and local improvement properties of mixing and deformation, then analyze displacement correction and prediction reuse.
Section~\ref{app:search-operators} shows that the search operations preserve node order, endpoints, and query budget, and gives local conditions for reducing the calibration loss.
Section~\ref{app:local-error} explains how correction strength and neighboring step sizes affect the leading local error, and relates the correction update to Adams--Bashforth.
Section~\ref{app:prefix-reuse} states when shared predictions can be reused without changing candidate outputs.

\subsection{Deformation and mixing}
\label{app:search-operators}
For a proposal $T$, let $\widetilde T$ be its schedule after prefix alignment to a fixed anchor $A$ (Appendix~\ref{app:candidates}).
For a fixed update rule, budget, and calibration set $\mathcal D_s$, define
\[
g(T)=\frac1n\sum_{i=1}^n L_i(\widetilde T),\qquad n=|\mathcal D_s|,
\]
where $L_i$ is the loss in Eq.~\ref{app:eq:loss} with fixed normalization constants $s_R,s_P,s_S$.
For the local analysis, node times are continuous and grid projection is omitted.

\textbf{Feasibility.}
For the suffix deformation in Eq.~\ref{app:eq:warp}, $W(0)=0$, $W(1)=1$, and
\begin{align}
W'(y)&=1+a(1-2y)+b(-6y^2+6y-1)\nonumber\\
&\geq 1-|a|-|b|=0.7>0,\qquad y\in[0,1].
\end{align}
Thus the suffix nodes remain strictly ordered.
The late-interval deformation preserves positive intervals and their total length by Eq.~\ref{app:eq:late-warp}.
For mixing, two schedules with common endpoints give
\begin{equation}
T_M=\frac{T_A+T_B}{2},\qquad
h_j^M=\frac{h_j^A+h_j^B}{2}>0.
\end{equation}
Both operations therefore preserve node order, endpoints, and budget.
The subsequent prefix alignment preserves these properties as well.

\textbf{Local decrease under deformation.}
Let $\Psi_T(a,b,\alpha)$ apply the suffix deformation followed by the late-interval deformation to $T$, and set $f=g\circ\Psi_T$, so $f(0)=g(T)$.
The six variants use parameter vectors $\pm d_1,\pm d_2,\pm d_3$, where
\begin{align}
d_1&=(0.15,0.15,0)^\top,\nonumber\\
d_2&=(0.15,-0.15,0)^\top,\qquad d_3=(0,0,0.35)^\top.
\label{app:eq:deformation-directions}
\end{align}
Assume $f$ is twice continuously differentiable near zero.
These directions span $\mathbb R^3$, so $\nabla f(0)\neq0$ implies $q_i=\nabla f(0)^\top d_i\neq0$ for some $i$.
Choose $s=-\operatorname{sign}(q_i)$.
Taylor expansion gives
\begin{equation}
f(s\epsilon d_i)-f(0)=-\epsilon|q_i|+O(\epsilon^2)<0
\label{app:eq:deformation-descent}
\end{equation}
for sufficiently small $\epsilon>0$.
Thus one of the six directions locally decreases the loss.
Our candidates use $\epsilon=1$; calibration determines which fixed-size deformations improve the output.

\textbf{Loss change under mixing.}
Let $\Delta T=T_A-T_B$ and $H=\nabla^2g(T_M)$, and assume $g$ is twice continuously differentiable near $T_M$.
For nearby schedules, Taylor expansion gives
\begin{align}
g(T_M\pm\Delta T/2)
={}&g(T_M)\pm\tfrac12\nabla g(T_M)^\top\Delta T\nonumber\\
&+\tfrac18\Delta T^\top H\Delta T+r_\pm,
\end{align}
where $r_\pm=o(\|\Delta T\|^2)$.
Averaging the two equations cancels the linear terms:
\begin{equation}
g(T_M)=\frac{g(T_A)+g(T_B)}2
-\frac18\Delta T^\top H\Delta T+R,
\label{app:eq:mixing-gain}
\end{equation}
with $R=-(r_++r_-)/2$.
Positive curvature along $\Delta T$ therefore lowers the midpoint loss relative to the mean seed loss, up to $R$.
The midpoint improves on both seeds whenever
\begin{equation}
\frac18\Delta T^\top H\Delta T-|R|
>\frac12|g(T_A)-g(T_B)|.
\label{app:eq:mixing-both}
\end{equation}
This condition compares the curvature gain with the difference between the two seed losses.

\subsection{Displacement correction and step size}
\label{app:local-error}
Let $x^*(t)$ solve Eq.~\ref{eq:ode} and define $w(t)=v_\theta(x^*(t),t;c)$, $w_j=w(t_j)$, and $\dot w_j=w'(t_j)$.
We use exact arithmetic and assume $w$ is twice continuously differentiable with $\|w''(t)\|\leq M_2$ on the current and preceding intervals.

\textbf{Local error.}
Taylor expansion gives
\begin{align}
x^*(t_{j+1})-x^*(t_j)
&=h_jw_j+\tfrac12h_j^2\dot w_j+R_x,\nonumber\\
w_j-w_{j-1}&=h_{j-1}\dot w_j+R_w,
\label{app:eq:taylor}
\end{align}
where $\|R_x\|\leq M_2h_j^3/6$ and $\|R_w\|\leq M_2h_{j-1}^2/2$.
Substituting these expansions into Eq.~\ref{eq:correction}, using exact current and preceding states, gives the local error (exact next state minus the corrected update):
\begin{align}
\tau_j&=\tfrac12(h_j^2-\eta h_{j-1}^2)\dot w_j+R_j,
\label{app:eq:local-defect}\\
\|R_j\|&\leq M_2\left(\frac{h_j^3}{6}+\frac{|\eta|h_{j-1}^3}{4}\right),\nonumber
\end{align}
where $R_j=R_x-\eta h_{j-1}R_w/2$.

\textbf{Correction strength.}
\label{app:damping}
For the step ratio $\rho_j=h_{j-1}/h_j$, correction multiplies Euler's leading error term by $1-\eta\rho_j^2$.
This multiplier has magnitude below one exactly when
\begin{equation}
|1-\eta\rho_j^2|<1
\quad\Longleftrightarrow\quad 0<\eta<\frac{2}{\rho_j^2}.
\label{app:eq:ratio-condition}
\end{equation}
For $\eta=0.5$, this requires $\rho_j<2$; with equal steps, the leading coefficient is half Euler's and the local error remains $O(h_j^2)$.
This range for $\eta$ depends on the neighboring step sizes.
Appendix~\ref{app:correction-strength} compares fixed strengths using generated outputs.

\textbf{Relation to Adams--Bashforth.}
Writing $p=x_j+h_ju_j$ for the Euler proposal, the variable-step AB2 proposal and our update satisfy
\begin{align}
p^{\rm AB2}&=p+\frac{h_j^2}{2h_{j-1}}(u_j-u_{j-1}),\nonumber\\
x_{j+1}&=(1-\eta\rho_j^2)p+\eta\rho_j^2p^{\rm AB2}.
\end{align}
For equal steps and $\eta=0.5$, our update averages the Euler and AB2 proposals from the same state and history.
On a nonuniform grid, the AB2 weight $\eta\rho_j^2$ can exceed one.

\subsection{Exact reuse of shared prefixes}
\label{app:prefix-reuse}
Fix the input, initialization, budget, update rule, and deterministic arithmetic.
Schedules sharing the first $q$ query times start from the same state and take the same steps, so their predictions along that prefix are identical.
For UniPC, the model time inputs and solver settings must also match, and each candidate maintains its own solver history.
The shared predictions can therefore be computed once and reused, after which each candidate follows its own remaining time nodes.
This reduces calibration cost without changing candidate outputs; the selected schedule still uses $B$ velocity evaluations for each new input.

\FloatBarrier
\section{Experimental protocol}
\label{app:protocol}
This section describes data use, reference samplers, metrics, and baseline settings for the experiments in Section~\ref{sec:experiments}.
Section~\ref{app:inputs} specifies the calibration and evaluation inputs, generation settings, and paired references.
Section~\ref{app:metrics} defines the image and video metrics.
Section~\ref{app:baseline-settings} gives the baseline settings and examines implementation choices through controlled comparisons.

\subsection{Inputs and references}
\label{app:inputs}
Table~\ref{tab:calibration_sets} summarizes the calibration and evaluation sets.
Table~\ref{tab:generation_settings} gives the common generation settings.
FLUX uses a maximum text length of 512 tokens.
Each guided Wan evaluation executes a conditional and an unconditional network branch.
For the trajectory-based image studies, $\mathcal D_s$ includes all 64 inputs in $\mathcal D_c$.
For Wan, $\mathcal D_c$ and $\mathcal D_s$ are disjoint.
Evaluation inputs are separate from both sets in every study.
Anchor selection uses the first four calibration inputs in $\mathcal D_c$ and screening uses the first eight.

\begin{table}[t!]
\centering
\apptablesetup
\footnotesize
\setlength{\tabcolsep}{2pt}
\captionsetup{position=top}
\caption{\textbf{Number of calibration and evaluation inputs.} We use $\mathcal D_c$ to construct seeds and screen candidates, and $\mathcal D_s$ to select the final schedules. ``Shared'' counts inputs included in both calibration sets. Evaluation inputs are separate from both sets.}
\label{tab:calibration_sets}
\begin{tabularx}{\columnwidth}{Xcccc}
\toprule[1pt]
\apphead{Study} & \apphead{$|\mathcal D_c|$} & \apphead{$|\mathcal D_s|$} & \apphead{Shared} & \apphead{Evaluation}\\
\midrule
Main images, $B=12,16$ & 64 & 128 & 64 & 500\\
Additional images, $B=8$ & 64 & 128 & 64 & 200\\
Additional images, $B=20$ & 64 & 64 & 64 & 200\\
Cross-style, $B=12,16$ & 64 & 128 & 64 & $4\times30$\\
Random seeds, $B=12$ & 8 & 128 & 8 & 200\\
Videos, $B=10,20$ & 24 & 24 & 0 & 100\\
\bottomrule[1pt]
\end{tabularx}
\end{table}

\begin{table}[t!]
\centering
\apptablesetup
\captionsetup{position=top}
\caption{\textbf{Image and video generation settings.} All runs use a batch size of one.}
\label{tab:generation_settings}
\begin{tabularx}{\columnwidth}{lXX}
\toprule[1pt]
\apphead{Setting} & \apphead{Images} & \apphead{Videos}\\
\midrule
Model & FLUX.1-dev & Wan2.2-TI2V-5B\\
Resolution & $1024\times1024$ & $832\times480$\\
Frames & 1 & 81 at 24 fps\\
Guidance scale & 3.5 & 5\\
Flow shift & Native FLUX & 5\\
Weight precision & BF16 & BF16\\
State precision & BF16 & FP32\\
Reference & Euler, 50 evaluations & UniPC, 50 guided evaluations\\
\bottomrule[1pt]
\end{tabularx}
\end{table}

Calibration and generation use the same sampler settings.
For FLUX, displacement correction subtracts the stored Euler increment from the trapezoidal displacement to account for BF16 rounding.
\TU{} uses Wan's flow-UniPC scheduler with order 2, \texttt{bh2}, and a lower-order final step.

FLUX-50 uses the native grid and 50 Euler evaluations; UniPC-50 uses Wan's native grid and 50 guided evaluations.
Each reference shares its conditioning input and base noise with the corresponding candidate output.
Cross-style calibration uses photography prompts only.
Its evaluation set contains 30 scenes in four styles, with the same base noise across styles and a separate FLUX-50 reference for each styled prompt.

\subsection{Metrics}
\label{app:metrics}
\textbf{Paired image fidelity.}
RMSE uses RGB values in $[0,255]$.
We compute LPIPS with AlexNet, scaling RGB pixel values to $[-1,1]$.
SSIM uses an $11\times11$ Gaussian window, standard deviation 1.5, population covariance, and range 255.
These metrics compare each generated image with its paired reference.

\textbf{FID.}
We compute \FIDref{} between the generated and FLUX-50 image sets for the same prompt--noise inputs, using 2,048-dimensional Inception-v3 features extracted with \texttt{torch-fidelity}.
All methods in each comparison use the same inputs and sample count.
Each set contains 500 images in the main experiment, 200 in the additional-budget and random-seed studies, and 30 per style, 120 across all four styles, or 90 across the three unseen styles in the cross-style generalization.
Pooled FID is recomputed from the pooled features.
FID is used only for evaluation; schedule selection uses RMSE, LPIPS, and SSIM.

\textbf{Human preference and text alignment.}
HPSv3 estimates human preference for each image--prompt pair; we use its predicted mean reward.
CLIP measures text alignment by the cosine similarity between image and text embeddings from ViT-L/14, with a maximum text length of 77 tokens.
Both scores are averaged over the evaluation inputs; higher values are better.

\textbf{Video metrics.}
For videos, PSNR uses all 81 frames; LPIPS and SSIM average 16 uniformly spaced frames from losslessly encoded outputs.
The six VBench dimensions are overall consistency, subject consistency, background consistency, motion smoothness, imaging quality, and dynamic degree.
They are reported separately, scaled by 100. Dynamic degree is the proportion of videos classified as non-static using optical flow.

\subsection{Baseline settings and controlled comparisons}
\label{app:baseline-settings}

We specify which baseline settings follow the original methods and which are adapted to our models and query budgets.
We use controlled comparisons to assess the effects of minor differences between our settings and those in the original papers.

GITS constructs schedules from 60-step iPNDM~\cite{zhang2023deis} trajectories on 256 calibration inputs and samples with Euler updates.
SharpEuler-Z uses native 50-step Euler trajectories on 64 calibration inputs and a density exponent of $0.5$.
SharpEuler-Wan uses native 50-step Euler trajectories on 24 calibration inputs with the same exponent; its schedules retain Wan's native initial time and end at zero noise.
OptimalSteps uses the authors' FLUX and Wan2.1 presets, applying the latter to Wan2.2-TI2V-5B.
It starts at $t_0$ from $(1-t_0)z$, where $z$ is the paired standard Gaussian noise.

CAT-OV uses moving-average coefficient $\beta=0.3$ and starts with two Euler steps of size $0.01$.
For each target budget $B$, we choose its step-size parameter $\lambda$ so that the mean query count over eight calibration inputs is within $5\%$ of $B$.
We then fix $\lambda$ for evaluation; query counts vary across inputs, and the tables report their observed mean.

For StableVS, we use the authors' UniPC-based implementation with their noise threshold $\sigma=0.85$ and no added noise.
For budget $B$, we start from a $\lceil1.5B\rceil$-interval UniPC grid with Wan's shift of 5.
We retain its high-noise query nodes ($\sigma\geq0.85$) and subsample the low-noise query nodes to obtain $B$ evaluations in total.
The multiplier $1.5$ is selected from $\{1.25,1.5\}$ by mean RMSE on eight calibration inputs for each budget.

\label{app:controls}
\begin{table*}[t!]
\centering
\apptablesetup
\captionsetup{position=top}
\caption{\textbf{Baseline settings and reference choice.} The OptimalSteps and SharpEuler comparisons change only the named setting and use FLUX-50 as the reference.
In the last block, we keep the generated outputs fixed and change the reference to iPNDM-60.}

\label{tab:supp_controls}
\label{tab:supp_reference}
\label{tab:supp_initialization}
\label{tab:supp_endpoint}
\resizebox{0.8\textwidth}{!}{%
\begin{tabularx}{\textwidth}{l*{6}{>{\centering\arraybackslash}X}}
\toprule[1pt]
& \multicolumn{3}{c}{\textbf{12 evaluations}} & \multicolumn{3}{c}{\textbf{16 evaluations}} \\
\cmidrule(lr){2-4}\cmidrule(l){5-7}
\apphead{Setting} & \apphead{RMSE$\downarrow$} & \apphead{LPIPS$\downarrow$} & \apphead{SSIM$\uparrow$} & \apphead{RMSE$\downarrow$} & \apphead{LPIPS$\downarrow$} & \apphead{SSIM$\uparrow$} \\
\midrule
\multicolumn{7}{@{}l}{\textit{OptimalSteps: initial-noise scaling (100 inputs; time nodes fixed)}} \\
Initial state $(1-t_0)z$ & 37.86 & 0.3804 & 0.7050 & 38.21 & 0.3542 & 0.7145 \\
Initial state $z$ & 37.01 & 0.3737 & 0.7111 & 37.16 & 0.3421 & 0.7236 \\
\midrule
\multicolumn{7}{@{}l}{\textit{SharpEuler: schedule settings (50 main images)}} \\
SharpEuler FLUX script & 34.28 & 0.5391 & 0.6455 & 23.74 & 0.3809 & 0.7426 \\
Full-range setting (SharpEuler-Z) & 34.35 & 0.3409 & 0.7164 & 22.81 & 0.2129 & 0.8028 \\
\midrule
\multicolumn{7}{@{}l}{\textit{SharpEuler: schedule settings (120 images across four styles)}} \\
SharpEuler FLUX script & 40.13 & 0.4819 & 0.6760 & 27.17 & 0.3306 & 0.7613 \\
Full-range setting (SharpEuler-Z) & 41.27 & 0.3374 & 0.7437 & 26.84 & 0.2017 & 0.8202 \\
\midrule
\multicolumn{7}{@{}l}{\textit{Reference changed to iPNDM-60 (100 inputs; generated outputs fixed)}} \\
GITS & 59.77 & 0.5747 & 0.5908 & --- & --- & --- \\
\TE & 33.19 & 0.3279 & 0.7365 & --- & --- & --- \\
\rowcolor{trajlight} \TC & 31.28 & 0.2837 & 0.7580 & --- & --- & --- \\
\specialrule{1pt}{0pt}{0pt}
\end{tabularx}%
}

\end{table*}

\textbf{OptimalSteps: initial-noise scaling.}
This control compares the authors' initialization $(1-t_0)z$ with unscaled noise $z$ (Table~\ref{tab:supp_controls}).
With OptimalSteps' time nodes fixed, changing its initial-noise scale to one lowers LPIPS from 0.3804 to 0.3737 at 12 evaluations and from 0.3542 to 0.3421 at 16.
Both TrajCal variants retain better RMSE, LPIPS, and SSIM at both budgets.

\textbf{SharpEuler: schedule settings.}
SharpEuler's general inverse-CDF construction reaches zero noise; its FLUX-specific variant retains part of the native low-noise grid.
The FLUX evaluation script ends at positive noise and replaces the penultimate node with a native node.
We compare this setting with the full-range construction, denoted SharpEuler-Z, keeping the calibrated density and Euler updates fixed.
We use SharpEuler-Z in the main comparisons to reach zero noise within the same budget $B$.
On 50 main images and 120 styled outputs, SharpEuler-Z gives lower LPIPS and higher SSIM at both budgets; 12-step main-image LPIPS falls from 0.5391 to 0.3409 (Table~\ref{tab:supp_controls}).
RMSE varies with the setting.

\textbf{Scoring reference: iPNDM-60.}
This comparison tests whether the fidelity advantage over GITS persists when the scoring reference uses its trajectory sampler.
GITS constructs its reference trajectories with 60 iPNDM evaluations.
We therefore rescore the existing 12-step outputs of GITS, \TE{}, and \TC{} against iPNDM-60 on 100 inputs, without reselecting any schedules.
GITS has LPIPS 0.5747, compared with 0.2837 for \TC{} and 0.3279 for \TE{}.
Both TrajCal variants also have lower RMSE and higher SSIM (Table~\ref{tab:supp_controls}).
Their advantage over GITS is therefore present under both scoring references.

\textbf{Additional baseline comparisons.}
\label{app:diagnostics}
We compare the following alternatives to our main baseline settings on eight calibration inputs at $B=12,16$.
The main GITS baseline retains the authors' default trajectory-deviation cost for schedule selection; L2 state error is evaluated here as an alternative.
Replacing the selection cost with L2 gives similar results at $B=12$; at $B=16$, RMSE decreases from 56.99 to 52.46, LPIPS from 0.5677 to 0.5154, and SSIM increases from 0.5763 to 0.6046.
For OptimalSteps, we compare BF16 time inputs and sampling states with two alternatives: FP32 time inputs only, or FP32 time inputs and states together, keeping the schedule, initialization, and network weights fixed.
Neither precision variant improves the three fidelity metrics.

\FloatBarrier
\section{Random and trajectory-based seed schedules}
\label{app:random-seeds}
This comparison tests whether TrajCal can start from randomly generated schedules of different shapes and whether trajectory-based seeds improve fidelity further.
We change the seed construction while retaining the same candidate expansion and selection procedure.
Both seed-construction variants use FLUX at 12 evaluations, the same 128 calibration inputs in $\mathcal D_s$ for final selection, and the same 200 evaluation inputs with FLUX-50 references.

Section~\ref{app:random-proposals} defines the random schedules and their parameter ranges.
Section~\ref{app:random-screen} describes how we select three seeds and apply the search procedure.
Section~\ref{app:random-results} compares output fidelity for random and trajectory-based seeds.

\subsection{Random shape proposals}
\label{app:random-proposals}
\begin{figure*}[t!]
\centering
\includegraphics[width=0.8\textwidth]{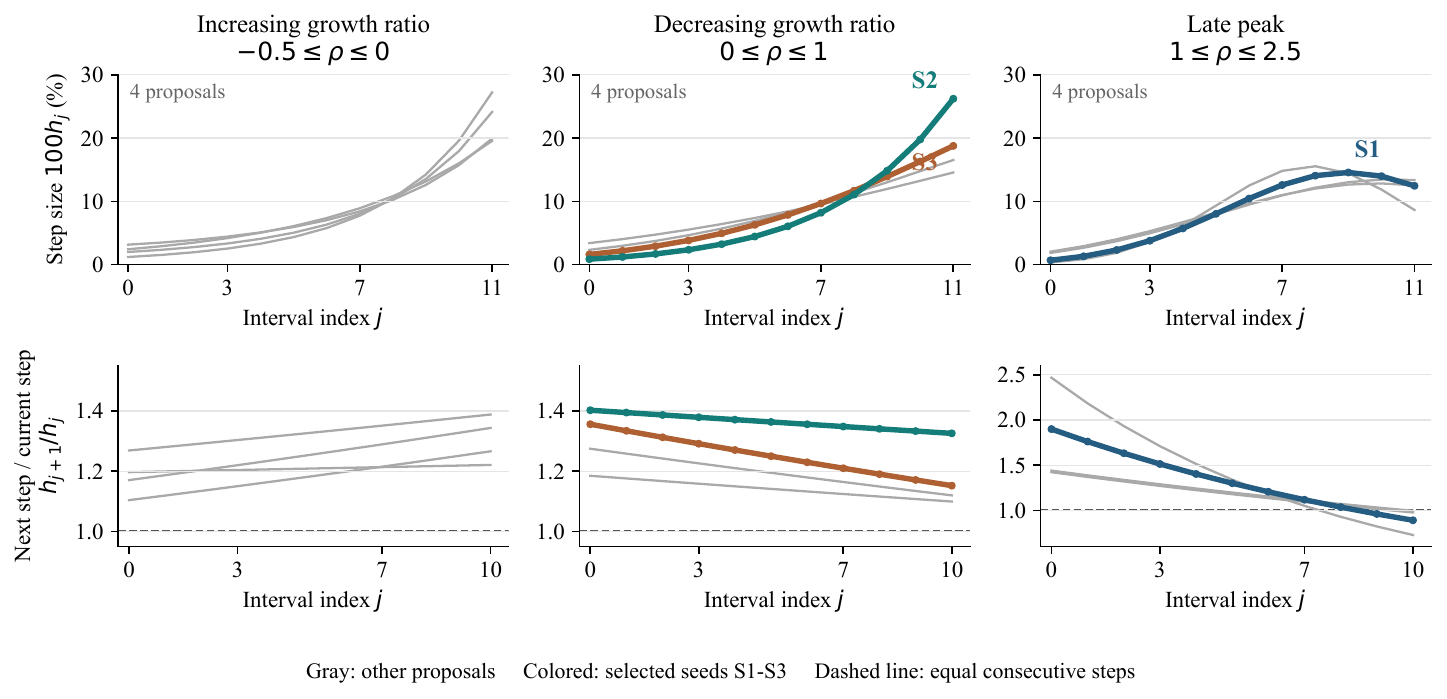}
\caption{\textbf{Random schedules and the three selected seeds.}
Each column shows four schedules sampled from one $\rho$ range at $B=12$.
Top: interval lengths $h_j$ as percentages of the full sampling interval.
Bottom: $h_{j+1}/h_j$, the ratio of the next interval length to the current one; the dashed line at one marks equal lengths.
Intervals grow throughout in the first two columns, but their growth ratios increase on the left and decrease in the middle.
In the right column, interval lengths peak near the end and then decrease.
Colors mark the selected seeds S1--S3; the remaining schedules are gray.
The schedules are shown before mixing, deformation, and prefix alignment.
Table~\ref{tab:random_parameters} gives all twelve parameter pairs.}
\label{fig:shape_priors}
\end{figure*}

\begin{table}[!htb]
\centering
\apptablesetup
\captionsetup{position=top}
\caption{\textbf{Parameters of the twelve random schedules.} The two indices in each name identify the $\rho$ range and $a$ subinterval. The last column gives the ratio of the last interval length to the first, $e^a=h_{11}/h_0$. Values are rounded for display.}
\label{tab:random_parameters}
\begin{tabularx}{\columnwidth}{l>{\raggedright\arraybackslash}Xrrr}
\toprule[1pt]
\apphead{$\rho$ range} & \apphead{Schedule} & \apphead{$a$} & \apphead{$\rho$} & \apphead{$e^a$}\\
\midrule
\multirow{4}{*}{$\rho\in[-0.5,0]$} & R1-1 & $1.8410$ & $-0.4501$ & $6.30$\\
 & R1-2 & $2.0849$ & $-0.0573$ & $8.04$\\
 & R1-3 & $2.4894$ & $-0.3352$ & $12.05$\\
 & R1-4 & $3.1125$ & $-0.1748$ & $22.48$\\
\midrule
\multirow{4}{*}{$\rho\in[0,1]$} & R2-1 & $1.4529$ & $0.3120$ & $4.28$\\
 & R2-2 & $1.9562$ & $0.4006$ & $7.07$\\
 & R2-3 & $2.4529$ & $0.4019$ & $11.62$\\
 & R2-4 & $3.4097$ & $0.0999$ & $30.25$\\
\midrule
\multirow{4}{*}{$\rho\in[1,2.5]$} & R3-1 & $1.8063$ & $1.2595$ & $6.09$\\
 & R3-2 & $1.9749$ & $1.1326$ & $7.21$\\
 & R3-3 & $2.8874$ & $1.5876$ & $17.95$\\
 & R3-4 & $3.2169$ & $2.2966$ & $24.95$\\
\bottomrule[1pt]
\end{tabularx}
\end{table}

\begin{table*}[t!]
\centering
\apptablesetup
\captionsetup{position=top}
\caption{\textbf{Random and trajectory-based seeds at 12 evaluations.} All methods use the same 200 image inputs and paired FLUX-50 references. Both seed constructions use the same candidate expansion and selection procedure. CAT-OV uses an adaptive number of evaluations. Bold and underlined values mark the best and second-best results, excluding the reference. Blue rows show \TC{}. HPSv3 and CLIP measure human preference and prompt alignment, respectively.}
\label{tab:random-seeds}
\resizebox{0.8\textwidth}{!}{%
\begin{tabularx}{\textwidth}{lc*{6}{>{\centering\arraybackslash}X}}
\toprule[1pt]
\multirow{2}{*}{\apphead{Method}} & \multirow{2}{*}{\apphead{Seed type}} & \multicolumn{4}{c}{\textbf{Reference fidelity}} & \multicolumn{2}{c}{\textbf{Quality}} \\
\cmidrule(lr){3-6}\cmidrule(lr){7-8}
 &  & \apphead{RMSE$\downarrow$} & \apphead{LPIPS$\downarrow$} & \apphead{SSIM$\uparrow$} & \apphead{FID$\downarrow$} & \apphead{HPSv3$\uparrow$} & \apphead{CLIP$\uparrow$} \\
\midrule
FLUX-12 & --- & 49.38 & 0.4754 & 0.6367 & 84.59 & 11.76 & \best{0.2664} \\
GITS & --- & 53.50 & 0.5381 & 0.6169 & 112.98 & 7.83 & 0.2611 \\
SharpEuler-Z & --- & 33.70 & 0.3286 & 0.7275 & 58.80 & 13.17 & \second{0.2659} \\
CAT-OV & --- & 36.34 & 0.3759 & 0.7017 & 76.67 & 11.68 & 0.2629 \\
OptimalSteps & --- & 35.67 & 0.3615 & 0.7176 & 68.65 & 12.44 & 0.2652 \\
\specialrule{0.6pt}{0pt}{2pt}
\TE{} & Random & 27.36 & 0.2700 & 0.7663 & 50.43 & \second{13.29} & 0.2650 \\
\rowcolor{trajlight} \TC{} & Random & 24.63 & 0.2271 & 0.7941 & \second{42.95} & \best{13.35} & 0.2641 \\
\specialrule{0.6pt}{0pt}{2pt}
\TE{} & Trajectory & \second{19.93} & \second{0.2103} & \second{0.8214} & 44.34 & 11.97 & 0.2643 \\
\rowcolor{trajlight} \TC{} & Trajectory & \best{19.75} & \best{0.1737} & \best{0.8413} & \best{40.70} & 12.12 & 0.2623 \\
\specialrule{0.6pt}{0pt}{2pt}
\rowcolor{refgray} FLUX-50 reference & --- & --- & --- & --- & --- & 12.20 & 0.2634 \\
\specialrule{1pt}{2pt}{0pt}
\end{tabularx}%
}
\end{table*}

Table~\ref{tab:random_parameters} lists all twelve parameter pairs.
For $j=0,\ldots,11$ and fixed $a,\rho$, let $q_j=j/11$ and define
\begin{equation}
h_j=\frac{\exp\!\left(a[q_j+\rho q_j(1-q_j)]\right)}
{\sum_{k=0}^{11}\exp\!\left(a[q_k+\rho q_k(1-q_k)]\right)}.
\label{app:eq:random-shapes}
\end{equation}
Cumulative sums of these positive intervals give nodes from zero to one.
The parameter $a$ sets the last-to-first interval ratio, $h_{11}/h_0=e^a$.
For this control, we choose a ratio range of $[4,32]$ to generate schedules with different degrees of concentration at early times, giving $a\in[\log4,\log32]$.
We divide this range into four equal subintervals and use three ranges for $\rho$: $[-0.5,0]$, $[0,1]$, and $[1,2.5]$.
Using random seed 20261112, we draw one parameter pair uniformly from each of the twelve combinations of an $a$ subinterval and a $\rho$ range.
In the first two groups, all sampled interval lengths increase, while the consecutive-interval ratio $h_{j+1}/h_j$ increases in the first group and decreases in the second.
The sampled proposals in the third group reach a late peak and then decrease.
All these schedules favor smaller early intervals, since $h_j/h_{11-j}=\exp[a(2q_j-1)]<1$ for $j=0,\ldots,5$.
Figure~\ref{fig:shape_priors} shows both interval lengths and consecutive-interval ratios, with the selected seeds highlighted.

\subsection{Selecting and expanding three seeds}
\label{app:random-screen}
We evaluate the twelve proposals on the eight calibration inputs in $\mathcal D_c$ under Euler and correction.
These eight inputs are included in the 128 calibration inputs in $\mathcal D_s$.
For each rule, we normalize RMSE, LPIPS, and $1-\mathrm{SSIM}$ as in Eq.~\ref{app:eq:scales}, and rank proposals by the combined loss averaged over inputs and rules.
Among the best eight, we retain the lowest-mean proposal and examine the 21 possible pairs from the other seven.
We choose the pair that minimizes the mean best-available loss of the resulting triple: take the smallest loss within the triple for each input and rule, then average.
Ties favor the largest minimum distance between log-interval vectors, followed by candidate name.
Thus output evaluation chooses a complementary triple from the randomly drawn shapes.
We select R3-3, R2-4, and R2-3 from Table~\ref{tab:random_parameters}, labeled S1, S2, and S3, respectively, in Figure~\ref{fig:shape_priors}.
This triple contains one proposal with a late peak and two with decreasing growth ratios.

\textbf{Applying the search.}
The three seeds enter the expansion procedure in Appendix~\ref{app:candidates}, giving 42 proposals with continuous nodes.
The first selected seed occupies the first construction slot, so its two suffix variants with $(a,b)=(-0.15,0.15)$ and $(0.15,0.15)$ are protected during screening.
Anchor selection uses the first four inputs in $\mathcal D_c$, screening uses all eight, and the 18 retained schedules are evaluated on all 128 inputs in $\mathcal D_s$.
Euler and correction select the same deformed version of the second random seed after prefix alignment.
Each method then reuses its selected schedule for 12 network evaluations on every evaluation input.

\subsection{Reference-fidelity results}
\label{app:random-results}
With random starts, both \TE{} and \TC{} improve all four fidelity measures over all five image baselines (Table~\ref{tab:random-seeds}).
For \TC{}, RMSE falls from SharpEuler-Z's 33.70 to 24.63, LPIPS from 0.3286 to 0.2271, and FID from 58.80 to 42.95, while SSIM rises from 0.7275 to 0.7941.
Thus the complete search already achieves strong fidelity with the randomly generated proposals.

Replacing the seeds with the three trajectory-based constructions further improves all four fidelity measures for both update rules.
For \TC{}, RMSE falls from 24.63 to 19.75 and LPIPS from 0.2271 to 0.1737, reductions of 19.8\% and 23.5\%.
FID decreases from 42.95 to 40.70 and SSIM increases from 0.7941 to 0.8413.
The expansion and selection procedure is unchanged.

The two random-start variants use identical time nodes, allowing a direct comparison of update rules.
Correction reduces RMSE by 10.0\% and LPIPS by 15.9\%, and raises SSIM from 0.7663 to 0.7941.
The random-start variants have higher HPSv3 scores than their trajectory-seeded counterparts.
These results show that candidate search is effective with randomly constructed seeds and that trajectory information provides stronger starting points.

\FloatBarrier
\section{Calibration choices and update-rule comparisons}
\label{app:design}
Sections~\ref{app:objective-control}--\ref{app:prefix-control} examine the selection objective, correction strength, and shared-prefix constraint at 12 evaluations (Table~\ref{tab:supp_ablations}).
Section~\ref{app:schedule-update} compares Euler and correction updates on fixed TrajCal schedules at 12 and 16 evaluations (Table~\ref{tab:ablation}).
The comparisons in this section use the first 100 caption--noise pairs from the 500-pair main image evaluation set.
These inputs are excluded from calibration.

\subsection{Selection objective}
\label{app:objective-control}
This comparison tests the combined fidelity loss against RMSE-only selection.
We hold the 18 retained candidates and 128 calibration inputs in $\mathcal D_s$ fixed, change only the selection objective, and evaluate the selected schedules on the same 100 image inputs at 12 evaluations.
\begin{table*}[t!]
\centering
\apptablesetup
\captionsetup{position=top}
\caption{\textbf{Selection objective, correction strength, and shared prefix.} We report mean scores on the same 100 image inputs at 12 evaluations.
The objective comparison uses the same retained candidates. We select a schedule separately for each $\eta$ and repeat screening and selection when prefix alignment is removed.}

\label{tab:supp_ablations}
\resizebox{0.8\textwidth}{!}{%
\begin{tabularx}{\textwidth}{l*{5}{>{\centering\arraybackslash}X}}
\toprule[1pt]
\apphead{Setting} & \apphead{RMSE$\downarrow$} & \apphead{LPIPS$\downarrow$} & \apphead{SSIM$\uparrow$} & \apphead{HPSv3$\uparrow$} & \apphead{CLIP$\uparrow$} \\
\midrule
\multicolumn{6}{@{}l}{\textit{Selection objective}} \\
\TE{}, RMSE only & 20.13 & 0.2171 & 0.8111 & 12.51 & 0.2702 \\
\TE{}, combined objective & 19.83 & 0.2061 & 0.8204 & 12.01 & 0.2696 \\
\TC{}, RMSE only & 20.23 & 0.1779 & 0.8356 & 12.19 & 0.2665 \\
\TC{}, combined objective & 20.23 & 0.1779 & 0.8356 & 12.19 & 0.2665 \\
\midrule
\multicolumn{6}{@{}l}{\textit{Correction strength in \TC{}}} \\
$\eta=0.25$ & 19.25 & 0.1786 & 0.8377 & 12.02 & 0.2672 \\
$\eta=0.5$ & 20.23 & 0.1779 & 0.8356 & 12.19 & 0.2665 \\
$\eta=1$ & 26.37 & 0.2272 & 0.7884 & 12.69 & 0.2683 \\
\midrule
\multicolumn{6}{@{}l}{\textit{Shared prefix in \TC{}}} \\
Only the initial time is shared & 21.04 & 0.1803 & 0.8321 & 12.42 & 0.2685 \\
Shared prefix (default) & 20.23 & 0.1779 & 0.8356 & 12.19 & 0.2665 \\
\bottomrule[1pt]
\end{tabularx}%
}

\end{table*}

\begin{table*}[t!]
\centering
\apptablesetup
\captionsetup{position=top}
\caption{\textbf{Euler and correction on fixed schedules.} We apply both updates to each selected schedule at 12 and 16 evaluations, using the same 100 image inputs. Time nodes are fixed within each comparison; correction uses $\eta=0.5$. Bold and underlined values mark the best and second-best results at each budget.}
\label{tab:ablation}
\resizebox{0.8\textwidth}{!}{%
\begin{tabularx}{\textwidth}{ll*{6}{>{\centering\arraybackslash}X}}
\toprule[1pt]
& & \multicolumn{3}{c}{\textbf{12 evaluations}} & \multicolumn{3}{c}{\textbf{16 evaluations}} \\
\cmidrule(lr){3-5}\cmidrule(l){6-8}
\apphead{Schedule selected for} & \apphead{Update rule} & \apphead{RMSE$\downarrow$} & \apphead{LPIPS$\downarrow$} & \apphead{SSIM$\uparrow$} & \apphead{RMSE$\downarrow$} & \apphead{LPIPS$\downarrow$} & \apphead{SSIM$\uparrow$} \\
\midrule
Euler & Euler & \best{19.83} & 0.2061 & 0.8204 & 21.52 & 0.1909 & 0.8246 \\
Euler & Correction & 20.88 & \second{0.1920} & \second{0.8256} & \best{17.87} & \second{0.1411} & \second{0.8615} \\
Correction & Euler & 20.69 & 0.2047 & 0.8204 & 22.23 & 0.1883 & 0.8253 \\
\rowcolor{trajlight} Correction & Correction & \second{20.23} & \best{0.1779} & \best{0.8356} & \second{18.47} & \best{0.1382} & \best{0.8640} \\
\specialrule{1pt}{0pt}{0pt}
\end{tabularx}%
}
\end{table*}

For \TE{}, the combined loss lowers RMSE from 20.13 to 19.83 and LPIPS from 0.2171 to 0.2061, and raises SSIM from 0.8111 to 0.8204 (Table~\ref{tab:supp_ablations}).
HPSv3 changes from 12.51 to 12.01.
For \TC{}, both objectives select the same schedule and give identical outputs.
These results support using the combined objective to select for pixel, perceptual, and structural fidelity together.

\subsection{Correction strength}
\label{app:ablations}
\label{app:correction-strength}
This comparison tests the correction coefficient in Eq.~\ref{eq:correction}.
We hold the 18 retained candidates fixed, evaluate each $\eta\in\{0.25,0.5,1\}$ on the same 128 calibration inputs in $\mathcal D_s$, and select one schedule per coefficient using the combined loss.
All three selected configurations are evaluated on the same 100 image inputs at 12 evaluations.

Both $\eta=0.25$ and $0.5$ improve RMSE, LPIPS, and SSIM over $\eta=1$ (Table~\ref{tab:supp_ablations}), supporting a partial correction.
The default $\eta=0.5$ gives RMSE 20.23, LPIPS 0.1779, and SSIM 0.8356, compared with 26.37, 0.2272, and 0.7884 for $\eta=1$.
The two partial corrections give similar LPIPS: 0.1786 for $\eta=0.25$ and 0.1779 for $\eta=0.5$, while $\eta=0.25$ gives lower RMSE and higher SSIM.
This comparison supports the default as an effective partial correction.

\subsection{Shared prefix}
\label{app:prefix-control}
This comparison tests whether aligning the initial query times affects reference fidelity.
We remove prefix alignment, retain only the common initial time, and repeat screening and final selection for \TC{} using the same calibration sets $\mathcal D_c$ and $\mathcal D_s$ and evaluation protocol.
At 12 evaluations on 100 images, this gives RMSE 21.04, LPIPS 0.1803, and SSIM 0.8321, compared with 20.23, 0.1779, and 0.8356 for the default procedure (Table~\ref{tab:supp_ablations}).
The shared-prefix constraint preserves comparable fidelity and reduces final candidate evaluation from 27,648 to 18,944 model calls per rule (Appendix~\ref{app:efficiency}).

\subsection{Update rule}
\label{app:schedule-update}
This comparison tests how displacement correction works with schedules selected by TrajCal (Table~\ref{tab:ablation}).
At each budget, we take the schedules selected for \TE{} and \TC{} and evaluate each with both Euler and correction updates on the same 100 paired images.
Time nodes remain fixed within each comparison; correction uses $\eta=0.5$ without additional model evaluations.

Correction lowers LPIPS and raises SSIM on both schedules at both budgets (Table~\ref{tab:ablation}).
On the \TE{} schedule, LPIPS changes from 0.2061 to 0.1920 at 12 evaluations and from 0.1909 to 0.1411 at 16.
The complete \TC{} configuration gives LPIPS 0.1779 and 0.1382, with SSIM 0.8356 and 0.8640, respectively.
RMSE depends on the schedule: at 12 evaluations correction raises it on the Euler-selected schedule but lowers it on the correction-selected schedule; at 16 it lowers RMSE on both.
These comparisons support displacement correction as an effective update within the TrajCal framework, particularly for perceptual and structural fidelity.

\FloatBarrier
\section{Additional evaluation}
\label{app:additional}
This section reports cross-style results, results at additional image budgets, and paired comparisons across inputs.
It then gives visual comparisons of video generation and image generation with caching methods.

\subsection{Cross-style generalization}
\label{app:styles}
This experiment tests whether schedules calibrated on photography transfer to unseen styles.
We keep each selected schedule fixed and change only the style prefix for the same 30 held-out scenes and base noise.
Every output has a FLUX-50 reference generated with the same styled prompt and base noise.

\textbf{Seed construction.}
For TrajCal's stepsize-search seed, we run OptimalSteps' state-relaxation search on the first four photography calibration inputs in $\mathcal D_c$, using the native 50-interval FLUX reference grid.
At each budget, we select one of the four resulting schedules by minimizing its mean absolute difference from the others at corresponding time nodes.
The other seed constructions and the remaining calibration steps follow Appendix~\ref{app:implementation}.


\begin{table}[!tb]
\centering
\apptablesetup
\footnotesize
\setlength{\tabcolsep}{1pt}
\renewcommand{\arraystretch}{1.06}
\captionsetup{position=top}
\caption{\textbf{Results for photography.} Calibration uses photography. Each budget uses the same 30 scenes and paired FLUX-50 references. Bold and underlined values mark the best and second-best results, excluding the reference.}
\label{tab:style-photography}
\label{tab:supp_styles}
\begin{tabularx}{\columnwidth}{@{}>{\raggedright\arraybackslash}X*{6}{c}@{}}
\toprule[1pt]
\apphead{Method} & \apphead{RMSE$\downarrow$} & \apphead{LPIPS$\downarrow$} & \apphead{SSIM$\uparrow$} & \apphead{FID$\downarrow$} & \apphead{HPSv3$\uparrow$} & \apphead{CLIP$\uparrow$} \\
\midrule
\multicolumn{7}{@{}l}{\textit{12 evaluations}} \\
FLUX-12 & 49.47 & 0.4967 & 0.6335 & 131.82 & 13.12 & 0.2836 \\
GITS & 52.29 & 0.5616 & 0.6144 & 152.10 & 10.37 & 0.2892 \\
SharpEuler-Z & 33.51 & 0.3320 & 0.7338 & 78.31 & \best{13.86} & 0.2874 \\
CAT-OV & 35.54 & 0.3874 & 0.7056 & 98.05 & 12.99 & 0.2881 \\
OptimalSteps & 30.62 & 0.3327 & 0.7449 & 83.12 & 12.86 & \second{0.2912} \\
\TE{} & \best{19.95} & \second{0.2315} & \second{0.8124} & \second{62.65} & \second{13.72} & \best{0.2926} \\
\rowcolor{trajlight} \TC{} & \second{22.38} & \best{0.2078} & \best{0.8193} & \best{56.10} & 13.67 & 0.2904 \\
\midrule
\multicolumn{7}{@{}l}{\textit{16 evaluations}} \\
FLUX-16 & 45.99 & 0.4526 & 0.6589 & 125.69 & 12.79 & 0.2896 \\
GITS & 52.41 & 0.5573 & 0.6096 & 147.30 & 10.88 & 0.2897 \\
SharpEuler-Z & 21.35 & 0.2041 & 0.8217 & 60.08 & \best{13.76} & 0.2891 \\
CAT-OV & 33.07 & 0.3256 & 0.7395 & 79.62 & 13.09 & 0.2903 \\
OptimalSteps & 30.02 & 0.3029 & 0.7558 & 75.81 & 13.50 & \second{0.2903} \\
\TE{} & \best{17.23} & \second{0.1688} & \second{0.8533} & \best{46.11} & \second{13.51} & 0.2885 \\
\rowcolor{trajlight} \TC{} & \second{19.96} & \best{0.1604} & \best{0.8537} & \second{48.12} & 13.45 & \best{0.2906} \\
\midrule
\rowcolor{refgray} FLUX-50 & --- & --- & --- & --- & 13.25 & 0.2946 \\
\bottomrule[1pt]
\end{tabularx}
\end{table}

\begin{table}[!tb]
\centering
\apptablesetup
\footnotesize
\setlength{\tabcolsep}{1pt}
\renewcommand{\arraystretch}{1.06}
\captionsetup{position=top}
\caption{\textbf{Results for watercolor.} Schedules calibrated on photography are used unchanged. Each budget uses the same 30 scenes and paired FLUX-50 references. Bold and underlined values mark the best and second-best results, excluding the reference.}
\label{tab:style-watercolor}
\begin{tabularx}{\columnwidth}{@{}>{\raggedright\arraybackslash}X*{6}{c}@{}}
\toprule[1pt]
\apphead{Method} & \apphead{RMSE$\downarrow$} & \apphead{LPIPS$\downarrow$} & \apphead{SSIM$\uparrow$} & \apphead{FID$\downarrow$} & \apphead{HPSv3$\uparrow$} & \apphead{CLIP$\uparrow$} \\
\midrule
\multicolumn{7}{@{}l}{\textit{12 evaluations}} \\
FLUX-12 & 59.74 & 0.5003 & 0.6642 & 171.69 & 9.49 & 0.2771 \\
GITS & 63.12 & 0.5650 & 0.6671 & 251.06 & 1.00 & 0.2696 \\
SharpEuler-Z & 46.12 & 0.3936 & 0.6883 & 114.81 & 12.57 & 0.2776 \\
CAT-OV & 43.52 & 0.3864 & 0.6981 & 122.91 & 11.41 & \best{0.2850} \\
OptimalSteps & 43.13 & 0.4211 & 0.6742 & 113.60 & \second{12.93} & \second{0.2849} \\
\TE{} & \second{25.22} & \second{0.2694} & \second{0.7537} & \second{72.97} & \best{13.44} & 0.2752 \\
\rowcolor{trajlight} \TC{} & \best{23.40} & \best{0.2021} & \best{0.7874} & \best{59.08} & 12.90 & 0.2762 \\
\midrule
\multicolumn{7}{@{}l}{\textit{16 evaluations}} \\
FLUX-16 & 56.78 & 0.4671 & 0.6715 & 162.36 & 10.51 & \second{0.2807} \\
GITS & 63.08 & 0.5608 & 0.6650 & 243.73 & 1.69 & 0.2693 \\
SharpEuler-Z & 30.12 & 0.2464 & 0.7668 & 71.41 & 12.37 & 0.2715 \\
CAT-OV & 42.85 & 0.3522 & 0.7149 & 114.32 & 11.57 & 0.2725 \\
OptimalSteps & 46.07 & 0.4306 & 0.6567 & 112.28 & \best{13.95} & \best{0.2829} \\
\TE{} & \second{22.44} & \second{0.2013} & \second{0.8017} & \second{57.89} & \second{12.38} & 0.2689 \\
\rowcolor{trajlight} \TC{} & \best{20.54} & \best{0.1403} & \best{0.8328} & \best{45.36} & 12.36 & 0.2726 \\
\midrule
\rowcolor{refgray} FLUX-50 & --- & --- & --- & --- & 12.08 & 0.2645 \\
\bottomrule[1pt]
\end{tabularx}
\end{table}

\begin{table}[!tb]
\centering
\apptablesetup
\footnotesize
\setlength{\tabcolsep}{1pt}
\renewcommand{\arraystretch}{1.06}
\captionsetup{position=top}
\caption{\textbf{Results for anime.} Schedules calibrated on photography are used unchanged. Each budget uses the same 30 scenes and paired FLUX-50 references. Bold and underlined values mark the best and second-best results, excluding the reference.}
\label{tab:style-anime}
\begin{tabularx}{\columnwidth}{@{}>{\raggedright\arraybackslash}X*{6}{c}@{}}
\toprule[1pt]
\apphead{Method} & \apphead{RMSE$\downarrow$} & \apphead{LPIPS$\downarrow$} & \apphead{SSIM$\uparrow$} & \apphead{FID$\downarrow$} & \apphead{HPSv3$\uparrow$} & \apphead{CLIP$\uparrow$} \\
\midrule
\multicolumn{7}{@{}l}{\textit{12 evaluations}} \\
FLUX-12 & 62.26 & 0.4896 & 0.6487 & 139.16 & 12.08 & \best{0.3275} \\
GITS & 66.73 & 0.5483 & 0.6466 & 194.30 & 7.80 & 0.3236 \\
SharpEuler-Z & 43.78 & 0.3284 & 0.7263 & 92.11 & 13.02 & 0.3263 \\
CAT-OV & 43.94 & 0.3482 & 0.7061 & 109.24 & 13.11 & 0.3253 \\
OptimalSteps & 49.98 & 0.4131 & 0.6787 & 129.61 & \best{13.37} & 0.3161 \\
\TE{} & \best{24.88} & \second{0.1915} & \second{0.8130} & \second{71.20} & 13.20 & 0.3234 \\
\rowcolor{trajlight} \TC{} & \second{27.57} & \best{0.1821} & \best{0.8149} & \best{67.31} & \second{13.29} & \second{0.3264} \\
\midrule
\multicolumn{7}{@{}l}{\textit{16 evaluations}} \\
FLUX-16 & 58.53 & 0.4444 & 0.6692 & 131.00 & 12.40 & 0.3273 \\
GITS & 66.82 & 0.5452 & 0.6416 & 186.04 & 8.37 & 0.3259 \\
SharpEuler-Z & 27.69 & 0.1755 & 0.8204 & \second{56.34} & 13.29 & \second{0.3302} \\
CAT-OV & 42.95 & 0.3168 & 0.7271 & 95.93 & 13.11 & 0.3273 \\
OptimalSteps & 50.07 & 0.3971 & 0.6847 & 117.35 & \best{13.49} & 0.3220 \\
\TE{} & \best{22.97} & \second{0.1501} & \second{0.8474} & 56.94 & 13.12 & 0.3252 \\
\rowcolor{trajlight} \TC{} & \second{24.72} & \best{0.1462} & \best{0.8479} & \best{54.00} & \second{13.31} & \best{0.3303} \\
\midrule
\rowcolor{refgray} FLUX-50 & --- & --- & --- & --- & 13.39 & 0.3313 \\
\bottomrule[1pt]
\end{tabularx}
\end{table}

\begin{table}[!tb]
\centering
\apptablesetup
\footnotesize
\setlength{\tabcolsep}{1pt}
\renewcommand{\arraystretch}{1.06}
\captionsetup{position=top}
\caption{\textbf{Results for flat-vector art.} Schedules calibrated on photography are used unchanged. Each budget uses the same 30 scenes and paired FLUX-50 references. Bold and underlined values mark the best and second-best results, excluding the reference.}
\label{tab:style-flat-vector}
\begin{tabularx}{\columnwidth}{@{}>{\raggedright\arraybackslash}X*{6}{c}@{}}
\toprule[1pt]
\apphead{Method} & \apphead{RMSE$\downarrow$} & \apphead{LPIPS$\downarrow$} & \apphead{SSIM$\uparrow$} & \apphead{FID$\downarrow$} & \apphead{HPSv3$\uparrow$} & \apphead{CLIP$\uparrow$} \\
\midrule
\multicolumn{7}{@{}l}{\textit{12 evaluations}} \\
FLUX-12 & 52.72 & 0.4387 & 0.8050 & 230.12 & 3.72 & 0.3041 \\
GITS & 56.15 & 0.4861 & 0.8096 & 313.28 & -4.03 & 0.2637 \\
SharpEuler-Z & 41.67 & 0.2956 & 0.8264 & 127.47 & 11.57 & 0.3197 \\
CAT-OV & 40.87 & 0.3007 & 0.8206 & 130.43 & 11.67 & 0.3187 \\
OptimalSteps & 46.26 & 0.3480 & 0.7948 & 148.11 & \best{12.58} & \second{0.3227} \\
\TE{} & \best{22.69} & \best{0.1383} & \best{0.8977} & \best{75.34} & 12.16 & 0.3207 \\
\rowcolor{trajlight} \TC{} & \second{24.46} & \second{0.1481} & \second{0.8902} & \second{76.80} & \second{12.31} & \best{0.3272} \\
\midrule
\multicolumn{7}{@{}l}{\textit{16 evaluations}} \\
FLUX-16 & 50.39 & 0.4021 & 0.8069 & 184.44 & 8.07 & 0.3131 \\
GITS & 56.13 & 0.4856 & 0.8085 & 310.82 & -3.75 & 0.2694 \\
SharpEuler-Z & 28.22 & 0.1807 & 0.8719 & 82.39 & \second{12.38} & 0.3240 \\
CAT-OV & 39.77 & 0.2846 & 0.8283 & 121.46 & 12.02 & 0.3174 \\
OptimalSteps & 47.93 & 0.3369 & 0.7986 & 126.52 & \best{12.59} & \second{0.3247} \\
\TE{} & \best{21.23} & \best{0.1171} & \best{0.9114} & \best{60.29} & 12.24 & 0.3240 \\
\rowcolor{trajlight} \TC{} & \second{23.80} & \second{0.1314} & \second{0.8977} & \second{72.36} & 12.30 & \best{0.3299} \\
\midrule
\rowcolor{refgray} FLUX-50 & --- & --- & --- & --- & 12.32 & 0.3310 \\
\bottomrule[1pt]
\end{tabularx}
\end{table}

Tables~\ref{tab:style-photography}--\ref{tab:style-flat-vector} report all six metrics separately for photography, watercolor, anime, and flat-vector art, with 30 paired scenes per style.
At both budgets, \TC{} improves RMSE, LPIPS, SSIM, and FID over all five baselines in every style.
On the three unseen styles, its LPIPS reductions relative to SharpEuler-Z range from 44.6\% to 49.9\% at 12 evaluations and from 16.7\% to 43.1\% at 16.
\TC{} also exceeds every baseline in CLIP similarity on flat-vector art at both budgets, although the highest HPSv3 scores come from baselines.

Across all four styles, \TC{} reduces pooled FID from 82.38 to 54.50 at 12 evaluations and from 56.59 to 46.26 at 16, relative to SharpEuler-Z (Table~\ref{tab:styles}).

\textbf{Combined results for unseen styles.}
\begin{table}[!t]
\centering
\apptablesetup
\footnotesize
\setlength{\tabcolsep}{2pt}
\captionsetup{position=top}
\caption{\textbf{Combined results for the three unseen styles.} Schedules calibrated on photography are applied unchanged to watercolor, anime, and flat-vector art. Each budget uses 30 scenes in three styles, giving 90 images per method. RMSE, LPIPS, and SSIM are averaged over these images; FID is computed jointly from all 90 outputs and their paired FLUX-50 references. Bold and underlined values mark the best and second-best results at each budget.}
\label{tab:unseen-style-fidelity}
\begin{tabularx}{\columnwidth}{@{}l*{4}{>{\centering\arraybackslash}X}@{}}
\toprule[1pt]
\apphead{Method} & \apphead{RMSE$\downarrow$} & \apphead{LPIPS$\downarrow$} & \apphead{SSIM$\uparrow$} & \apphead{FID$\downarrow$} \\
\midrule
\multicolumn{5}{@{}l}{\textit{12 evaluations}} \\
Native Euler & 58.24 & 0.4762 & 0.7060 & 143.04 \\
GITS & 62.00 & 0.5331 & 0.7078 & 200.20 \\
SharpEuler-Z & 43.86 & 0.3392 & 0.7470 & 93.73 \\
CAT-OV & 42.78 & 0.3451 & 0.7416 & 101.02 \\
OptimalSteps & 46.46 & 0.3941 & 0.7159 & 105.84 \\
\TE{} & \best{24.26} & \second{0.1997} & \second{0.8214} & \second{64.01} \\
\rowcolor{trajlight} \TC{} & \second{25.14} & \best{0.1774} & \best{0.8309} & \best{59.12} \\
\midrule
\multicolumn{5}{@{}l}{\textit{16 evaluations}} \\
Native Euler & 55.23 & 0.4379 & 0.7159 & 126.57 \\
GITS & 62.01 & 0.5305 & 0.7050 & 196.00 \\
SharpEuler-Z & 28.68 & 0.2009 & 0.8197 & 61.06 \\
CAT-OV & 41.86 & 0.3178 & 0.7568 & 92.93 \\
OptimalSteps & 48.02 & 0.3882 & 0.7133 & 97.36 \\
\TE{} & \best{22.21} & \second{0.1562} & \second{0.8535} & \second{51.42} \\
\rowcolor{trajlight} \TC{} & \second{23.02} & \best{0.1393} & \best{0.8595} & \best{50.03} \\
\bottomrule[1pt]
\end{tabularx}
\end{table}

Table~\ref{tab:unseen-style-fidelity} combines the results for watercolor, anime, and flat-vector art, excluding photography.
Each method contributes 90 outputs from the same 30 scenes.
We average the three paired metrics over these outputs and recompute FID from their pooled features and the corresponding FLUX-50 features.
\TC{} improves all four measures over every baseline at both budgets.
Relative to SharpEuler-Z, FID falls from 93.73 to 59.12 at 12 evaluations and from 61.06 to 50.03 at 16.

\subsection{Additional image budgets}
\label{app:budgets}
This comparison tests \TC{} at budgets below and above the main image settings.
We calibrate its schedules for 8 and 20 evaluations using the same procedure and compare it with the five baselines on 200 paired FLUX inputs, using FLUX-50 references.
\begin{table*}[t!]
\centering
\apptablesetup
\captionsetup{position=top}
\caption{\textbf{Image generation at 8 and 20 evaluations.} Both budgets use the same 200 image inputs and paired FLUX-50 references. CAT-OV averages 7.58 and 19.41 evaluations, respectively. Bold and underlined values mark the best and second-best results at each budget.}

\label{tab:supp_budgets}
\resizebox{0.8\textwidth}{!}{%
\begin{tabularx}{\textwidth}{l*{8}{>{\centering\arraybackslash}X}}
\toprule[1pt]
& \multicolumn{4}{c}{\textbf{8 evaluations}} & \multicolumn{4}{c}{\textbf{20 evaluations}} \\
\cmidrule(lr){2-5}\cmidrule(l){6-9}
\apphead{Method} & \apphead{RMSE$\downarrow$} & \apphead{LPIPS$\downarrow$} & \apphead{SSIM$\uparrow$} & \apphead{\FIDref{}$\downarrow$} & \apphead{RMSE$\downarrow$} & \apphead{LPIPS$\downarrow$} & \apphead{SSIM$\uparrow$} & \apphead{\FIDref{}$\downarrow$} \\
\midrule
Native Euler & 53.73 & 0.5410 & 0.6011 & 97.96 & 41.80 & 0.3819 & 0.6944 & 71.60 \\
GITS & 54.52 & 0.5696 & 0.6145 & 130.99 & 46.23 & 0.4457 & 0.6583 & 87.87 \\
SharpEuler-Z & 40.80 & 0.4258 & 0.6661 & 74.68 & \second{19.80} & \second{0.1634} & 0.8455 & \second{34.34} \\
CAT-OV & 39.23 & 0.4946 & 0.6552 & 120.09 & 31.77 & 0.2817 & 0.7597 & 55.88 \\
OptimalSteps & 33.33 & 0.3881 & 0.7019 & 77.19 & 38.76 & 0.3533 & 0.7134 & 66.40 \\
\TE{} & \second{31.31} & \second{0.3616} & \second{0.7148} & \second{68.24} & 20.64 & 0.1636 & \second{0.8486} & 35.11 \\
\rowcolor{trajlight} \TC{} & \best{29.30} & \best{0.3308} & \best{0.7322} & \best{64.04} & \best{17.15} & \best{0.1208} & \best{0.8819} & \best{28.22} \\
\bottomrule[1pt]
\end{tabularx}%
}

\end{table*}

Table~\ref{tab:supp_budgets} reports the results.
Both budgets use 64 calibration inputs in $\mathcal D_c$ for seed construction; $\mathcal D_s$ contains 128 inputs at budget 8 and 64 at budget 20.
\TC{} improves all four reference-fidelity metrics over all five baselines at both budgets.
At 8 evaluations, it reduces LPIPS by 14.8\% relative to OptimalSteps and achieves FID 64.04, the lowest among the compared samplers.
At 20, it reduces LPIPS by 26.1\% relative to SharpEuler-Z and lowers FID from 34.34 to 28.22.
\TC{} also improves all four fidelity metrics over \TE{} at both budgets; at 20 evaluations, LPIPS decreases from 0.1636 to 0.1208 and FID from 35.11 to 28.22.
\TC{} has HPSv3 scores of 12.93 and 12.24 at the two budgets, compared with 12.20 for FLUX-50 on this cohort.

\subsection{Paired comparisons across inputs}
\label{app:paired-results}
\label{app:statistics}
We analyze the existing evaluation outputs to check how often TrajCal improves fidelity on individual inputs and to estimate uncertainty in the mean gains.
The comparisons are \TC{} versus SharpEuler-Z for images and unseen styles, and \TU{} versus OptimalSteps for videos.
Each comparison uses the same prompt, base noise, and reference for both methods.

\textbf{Paired analysis.}
For each image or video input, we compute baseline minus TrajCal for RMSE/LPIPS, and TrajCal minus baseline for PSNR/SSIM; positive values therefore indicate improvement.
We count inputs where all three fidelity metrics improve.
For the mean improvement, we draw 10,000 bootstrap samples of the original sample size, with replacement and with the same input indices for both methods.
For each resample, we recompute the mean improvement; the 2.5th and 97.5th percentiles across resamples give the 95\% confidence interval.
Schedules and sampler settings remain fixed.

\textbf{Images.}
At equal budgets of 12 and 16 evaluations, \TC{} improves mean RMSE, LPIPS, and SSIM over SharpEuler-Z on 500 inputs, with all 95\% improvement intervals above zero.
All three metrics improve simultaneously on 454/500 and 384/500 inputs, respectively.
For 12-evaluation \TC{} versus 16-evaluation SharpEuler-Z, the mean LPIPS reduction is 0.0256, with a 95\% interval of $[0.0180,0.0336]$.
All three mean-improvement intervals remain above zero; all three metrics improve together on 247/500 inputs.

\textbf{Videos.}
At equal budgets of 10 and 20 evaluations, \TU{} improves PSNR, LPIPS, and SSIM over OptimalSteps on all 100 videos, with all three mean-improvement intervals above zero.
For 10-evaluation \TU{} versus 20-evaluation OptimalSteps, the mean LPIPS reduction is 0.1219, with a 95\% interval of $[0.1060,0.1380]$.
All three mean-improvement intervals remain above zero, and all three metrics improve simultaneously on 89/100 inputs.

\textbf{Unseen styles.}
For each method, we first average LPIPS over watercolor, anime, and flat-vector art within each of the 30 scenes, then apply the same paired bootstrap to these scene averages.
For each resample, the relative reduction is one minus the ratio of \TC{}'s mean LPIPS to SharpEuler-Z's.
The overall reductions are 47.7\% at 12 evaluations and 30.7\% at 16, with 95\% intervals of $[43.8\%,51.3\%]$ and $[23.3\%,37.8\%]$, respectively.
Scene-average LPIPS improves on 30/30 and 27/30 scenes, respectively.

\clearpage
\subsection{Visual comparison for video generation}
\label{app:video-qualitative}
\begin{figure*}[t!]
\centering
\includegraphics[width=\textwidth]{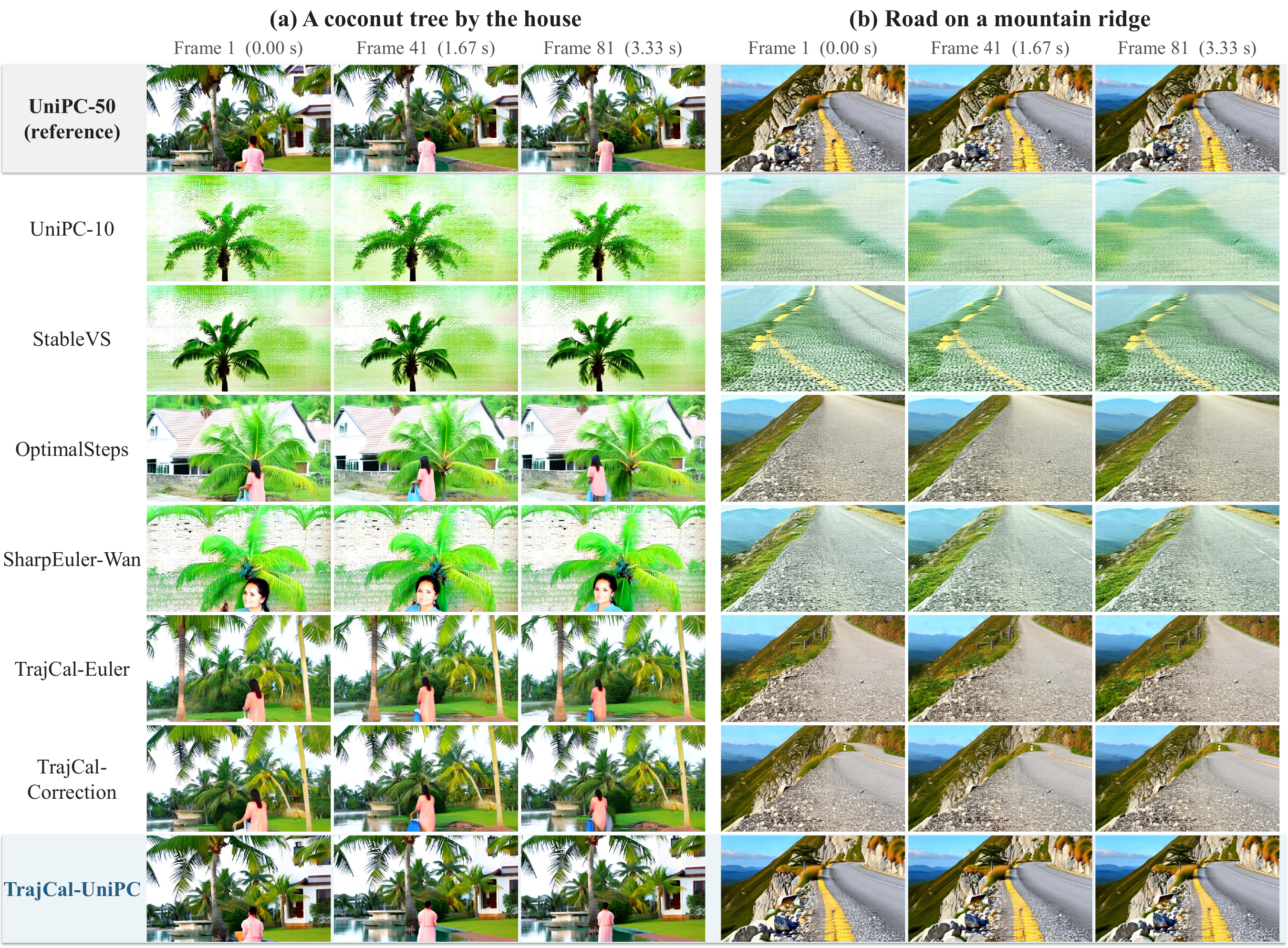}
\caption{\textbf{Video generation at 10 guided evaluations.}
The two examples use the prompts ``a coconut tree by the house'' and ``road on mountain ridge.''
Each example shows frames 1, 41, and 81 of an 81-frame video at 24 fps. The same frames are shown for every method.
The top row is the UniPC-50 reference; all other rows use 10 guided evaluations with the same prompt and base noise.
\TU{} more closely preserves the reference scene layout and appearance, including the pool and house in (a) and the road markings and rock face in (b).}
\label{fig:video_comparison}
\end{figure*}

This comparison examines how well ten-evaluation video outputs preserve the content and appearance of their UniPC-50 references.
Figure~\ref{fig:video_comparison} uses two held-out prompts and compares the four video baselines
\newpage\noindent
with the three TrajCal variants.

Within each example, all methods share the prompt and base noise, and frames are displayed at the same temporal positions.

\clearpage
\subsection{Visual comparison with caching methods}
\label{app:cache}
\begin{figure*}[t!]
\centering
\includegraphics[width=0.8\textwidth]{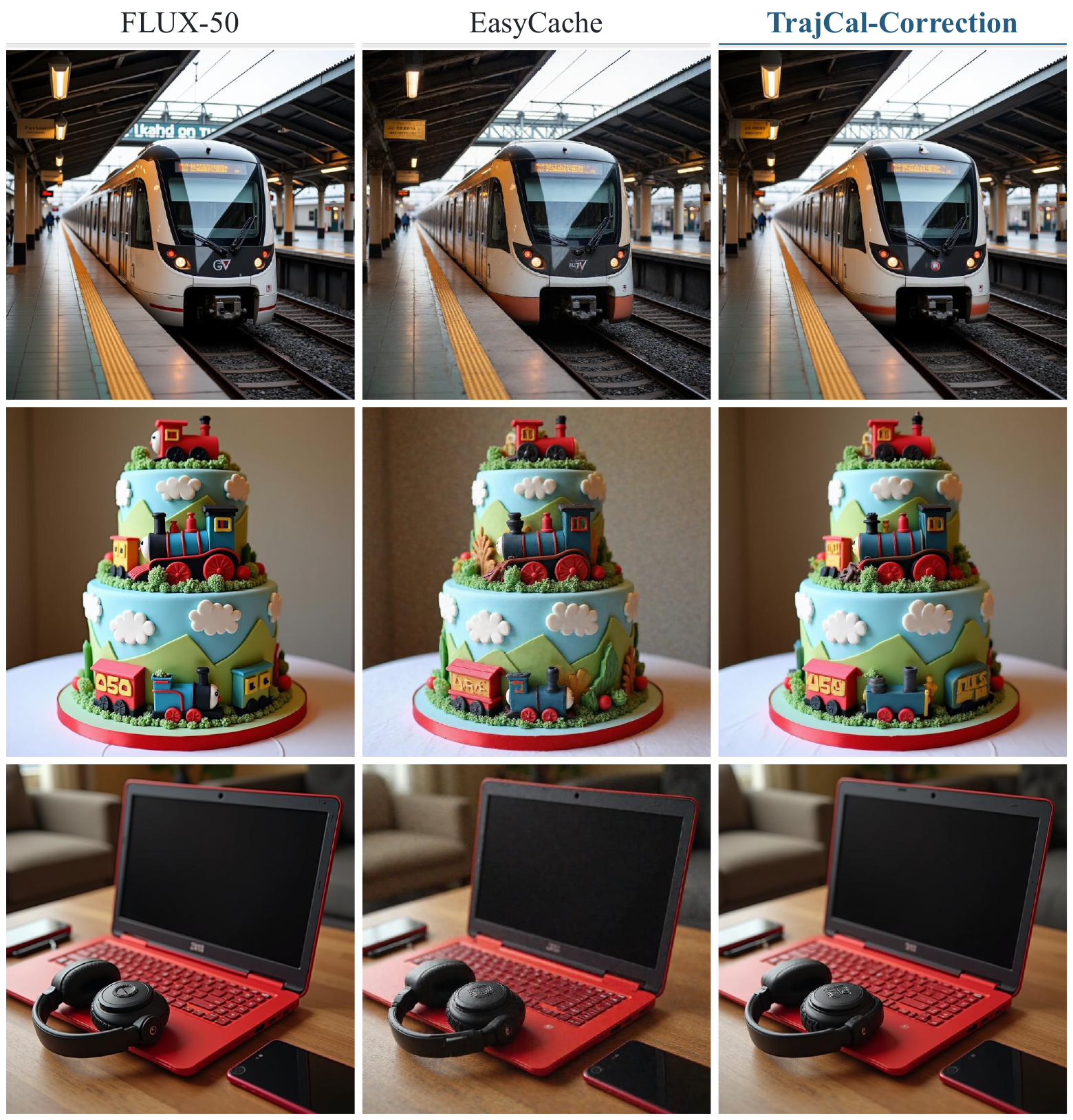}
\caption{\textbf{Image generation with EasyCache and \TC{}.} Each row uses one prompt--noise pair. From left to right: the FLUX-50 reference, EasyCache with 13 full model evaluations, and \TC{} with 12. EasyCache preserves the overall reference composition but leaves visible noise, particularly in the cake background and laptop screen; \TC{} produces cleaner outputs.}
\label{fig:cache_qualitative}
\end{figure*}

Caching accelerates sampling by reusing results from earlier network evaluations to approximate predictions at later steps.
EasyCache~\cite{zhou2025easycache} uses an error estimate to decide when to reuse cached results and when to evaluate the full network.

This comparison examines visible artifacts when caching and few-step sampling use similar numbers of full network evaluations.
Figure~\ref{fig:cache_qualitative} compares \TC{} with EasyCache
on three paired FLUX inputs, using the same prompts and base noise as the FLUX-50 references.
EasyCache retains 50 update positions but performs 13 full network evaluations in each example, while \TC{} uses 12.

EasyCache preserves the main objects and overall composition of the references, but visible noise remains, especially in the cake background and the laptop screen.
\TC{} retains these structures with cleaner surfaces and backgrounds.
Similar residual noise is also observed in cached video outputs.

\clearpage
\setlength{\intextsep}{6pt}
\section{Generation time and prefix-reuse savings}
\label{app:efficiency}
We report image and video generation times and candidate-evaluation calls saved by shared prefixes.
Timings use an NVIDIA RTX PRO 6000 with 96\,GB memory.
We report recorded denoising time, excluding model loading, prompt encoding, decoding, file writing, and scoring.

\subsection{Image generation time}
\label{app:costs}
The image panel of Table~\ref{tab:generation_time} reports all seven methods on the same 500 evaluation inputs at budgets 12 and 16.
The fixed-budget methods have similar mean times: approximately 5.1 and 6.8 seconds per image, respectively.
CAT-OV uses fewer evaluations on average, 11.57 and 15.50, and has correspondingly shorter mean times.
\begin{table*}[t!]
\centering
\apptablesetup
\captionsetup{position=top}
\caption{\textbf{Image and video generation time.} Mean $\pm$ sample standard deviation of recorded denoising time in seconds, on the same 500 image inputs or 100 video inputs used in the main comparisons. Budgets count velocity evaluations, with guidance included for video. CAT-OV uses 11.57 and 15.50 evaluations on average at the two image budgets. Decoding is excluded.}
\label{tab:generation_time}
\resizebox{0.8\textwidth}{!}{%
\begin{tabularx}{\textwidth}{l*{2}{>{\centering\arraybackslash}X}@{\hspace{12pt}}l*{2}{>{\centering\arraybackslash}X}}
\toprule[1pt]
\multicolumn{3}{c}{\textbf{Image generation (FLUX)}} & \multicolumn{3}{c}{\textbf{Video generation (Wan)}} \\
\cmidrule(lr){1-3}\cmidrule(l){4-6}
\apphead{Method} & \apphead{12 evaluations} & \apphead{16 evaluations} & \apphead{Method} & \apphead{10 evaluations} & \apphead{20 evaluations} \\
\midrule
FLUX & 5.115 $\pm$ 0.012 & 6.826 $\pm$ 0.016 & UniPC & 13.959 $\pm$ 0.010 & 27.922 $\pm$ 0.018 \\
GITS & 5.113 $\pm$ 0.027 & 6.825 $\pm$ 0.021 & StableVS & 13.943 $\pm$ 0.013 & 27.933 $\pm$ 0.019 \\
SharpEuler-Z & 5.116 $\pm$ 0.014 & 6.828 $\pm$ 0.020 & OptimalSteps & 13.932 $\pm$ 0.012 & 27.905 $\pm$ 0.019 \\
CAT-OV & 4.952 $\pm$ 0.280 & 6.645 $\pm$ 0.339 & SharpEuler-Wan & 13.928 $\pm$ 0.044 & 27.886 $\pm$ 0.091 \\
OptimalSteps & 5.114 $\pm$ 0.019 & 6.828 $\pm$ 0.012 & \TE{} & 13.942 $\pm$ 0.017 & 27.903 $\pm$ 0.017 \\
\TE{} & 5.118 $\pm$ 0.011 & 6.836 $\pm$ 0.016 & \TC{} & 13.938 $\pm$ 0.009 & 27.905 $\pm$ 0.019 \\
\rowcolor{trajlight} \TC{} & 5.116 $\pm$ 0.010 & 6.830 $\pm$ 0.016 & \TU{} & 13.920 $\pm$ 0.012 & 27.869 $\pm$ 0.018 \\
\bottomrule[1pt]
\end{tabularx}%
}
\end{table*}

\subsection{Video generation time}
The video panel of Table~\ref{tab:generation_time} reports all seven methods on the same 100 evaluation inputs at budgets 10 and 20.
Mean times are similar across methods at each budget.
For \TU{}, denoising takes 13.920 and 27.869 seconds per video on average, respectively.

\subsection{Shared-prefix call savings}
\label{app:calibration-cost}
Table~\ref{tab:image_calibration_calls} reports model calls during image calibration, separately for each update rule.
Let $K$ be the number of candidate schedules, $n$ the number of calibration inputs, and $B$ the calls per candidate and input.
Evaluating every candidate independently requires $KnB$ calls.

All candidates share their first $q=\max(2,\lceil B/4\rceil)$ query times, including $t_0$.
If we reuse only these $q$ predictions, the total is
\[
n\bigl[q+K(B-q)\bigr]
\]
calls: $nq$ for the shared predictions and $nK(B-q)$ for the remaining queries.
Some candidates share more than $q$ initial query times, allowing further reuse.
The table includes these additional savings.
At $B=12$, screening uses 2,456 instead of 4,032 calls, and final selection uses 18,944 instead of 27,648, a 31.5\% reduction for final selection.

\begin{table}[!ht]
\centering
\apptablesetup
\footnotesize
\captionsetup{position=top}
\caption{\textbf{Model calls with and without prefix reuse.} Counts are per update rule for image calibration. Screening uses 42 candidates and eight inputs; final selection uses 18 candidates and 128 inputs (64 at budget 20). Each stage starts with an empty cache. Counts exclude reference and anchor generation, decoding, and metric computation.}
\label{tab:image_calibration_calls}
\begin{tabularx}{\columnwidth}{c*{4}{>{\centering\arraybackslash}X}}
\toprule[1pt]
& \multicolumn{2}{c}{\textbf{Screening}} & \multicolumn{2}{c}{\textbf{Final selection}} \\
\cmidrule(lr){2-3}\cmidrule(l){4-5}
\apphead{Budget} & \apphead{No reuse} & \apphead{With reuse} & \apphead{No reuse} & \apphead{With reuse} \\
\midrule
8 & 2,688 & 1,624 & 18,432 & 11,648 \\
12 & 4,032 & 2,456 & 27,648 & 18,944 \\
16 & 5,376 & 3,208 & 36,864 & 25,088 \\
20 & 6,720 & 4,000 & 23,040 & 15,360 \\
\bottomrule[1pt]
\end{tabularx}
\end{table}

\end{document}